\pdfoutput=1

\documentclass[11pt]{article}

\usepackage{EMNLP2022}
\usepackage{times}
\usepackage{latexsym}

\usepackage[T1]{fontenc}

\usepackage[utf8]{inputenc}

\usepackage{microtype}

\usepackage{inconsolata}

\usepackage{multirow}
\usepackage{makecell}

\usepackage{pgfplots}
\pgfplotsset{compat=1.17}
\usepackage{subcaption}

\usepackage{amsmath,amsfonts,amssymb}
\usepackage{graphicx}

\usepackage{kotex}

\title{LittleBird: Efficient Faster \& Longer Transformer for Question Answering}

\author{Minchul Lee \and Kijong Han \and Myeong Cheol Shin \\
  Kakao Enterprise Corp., South Korea \\
  \texttt{\{phil.i,mat.h,index.sh\}@kakaoenterprise.com}
  }

\begin{document}
\maketitle
\begin{abstract}
BERT has shown a lot of sucess in a wide variety of NLP tasks. But it has a limitation dealing with long inputs due to its attention mechanism. Longformer, ETC and BigBird addressed this issue and effectively solved the quadratic dependency problem.
However we find that these models are not sufficient, and propose LittleBird, a novel model based on BigBird with improved speed and memory footprint while maintaining accuracy.
In particular, we devise a more flexible and efficient position representation method based on Attention with Linear Biases (ALiBi). We also show that replacing the method of global information represented in the BigBird with pack and unpack attention is more effective.
The proposed model can work on long inputs even after being pre-trained on short inputs, and can be trained efficiently reusing existing pre-trained language model for short inputs. This is a significant benefit for low-resource languages where large amounts of long text data are difficult to obtain.
As a result, our experiments show that LittleBird works very well in a variety of languages, achieving high performance in question answering tasks, particularly in KorQuAD2.0, Korean Question Answering Dataset for long paragraphs.
\end{abstract}

\section{Introduction}
Transformer~\citep{vaswani2017attention} and pre-trained language models~\citep{devlin2018bert, liu2019roberta} based on it have shown a lot of success in a wide variety of NLP tasks. However, the quadratic dependency problem that comes from the attention mechanism makes it impractical to process long documents. Many techniques have been studied to overcome this problem and BigBird~\citep{zaheer2020big} showed robust and state-of-the-art performance on various NLP downstream tasks.

In this study, we propose a new model LittleBird by analyzing and improving the shortcomings of BigBird. LittleBird shows improved speed and memory footprint compared to BigBird while maintaining the overall accuracy of the question answering (QA) benchmarks and showing better accuracy in some of them.

In this study, we propose three major improvements compared to BigBird. The first is the method for position representation. In BigBird, trainable positional embedding is used similar to BERT~\citep{devlin2018bert}, and in ETC, relative positional encoding is used similar to T5~\citep{raffel2020exploring}. However, trainable positional embedding cannot handle longer inputs than those used for training and the relative position encoding is relatively slow and uses extra memory and parameters~\citep{ma2021luna}. \citet{press2021train} introduced the attention with linear biases (ALiBi) method that resolves these problems, but it was designed for causal language modeling, not autoencoding language modeling, which is typically useful for QA tasks. Thus, we devise a new method based on the ALiBi that is fast, flexible, and also effective in QA tasks.

The second is the method of capturing global information. BigBird introduces two ways of capturing global information, the random sparse attention and global tokens~\citep{ainslie2020etc} which attend to and be attended by all other tokens. However, the random attention method is practically slow compared to its time complexity because it requires to repeat gather and scatter operations at random positions. In addition, a relatively large number of ($\sim$hundreds) global tokens are required to achieve the reported performance using only global tokens without a random attention method in ETC. We show that replacing them with modified pack and unpack attention~\citep{ma2021luna} is more effective. 

The last is the efficient way to train a model for long sequences. We introduce a simple but effective method, Padding Insertion, which makes the model robust to long inputs while training on short inputs. We also propose a distillation method that can maximize the reuse of the pre-trained model for a short length and show that our model can be effectively pre-trained using these methods.

Our model shows a 12$\sim$29\% reduction in peak memory usage and a 6$\sim$46\% reduction in latency compared to various BigBird and ETC model settings reported in the paper for 4K length document inference while showing better accuracy on several English QA benchmarks dev sets~\citep{kwiatkowski2019natural, welbl2018constructing}. Our model achieves new state-of-the-art performance on KorQUAD 2.0~\citep{kim2019korquad}, a Korean long document QA benchmark. In addition, these results are obtained with LittileBird pre-trained with only 2K sequence length. It shows that our novel positional representation method works well when the model is applied to the QA downstream task with a document longer than the sequence used in the pre-training phase.

\section{Background and Related Work}

\subsection{Transformers for Longer Input}
Various methods have been studied to maintain reasonable performance without using the quadratic operation of Transformer to handle long documents. \citet{child2019generating} introduced sparse factorizations of the attention matrix which reduce the complexity to $O(n\sqrt{n})$. Reformer~\citep{kitaev2019reformer} reduced complexity to $O(n\log{n})$ using locality-sensitive hashing. 

Longformer~\citep{beltagy2020longformer} and ETC~\citep{ainslie2020etc} proposed a method that utilizes several global attention tokens and local windowed attention and reduced the complexity to $O(n)$. In addition, these works showed performance improvement in downstream tasks. BigBird~\citep{zaheer2020big}, an extended study related to ETC, propose random sparse attention method and provides detailed theoretical background and experimental results for more downstream tasks.

Recently, LUNA~\citep{ma2021luna} introduced the method that approximates softmax attention with two nested linear attention functions called Pack and Unpack Attention, which has only linear time complexity. This method recorded improved performance in both speed and score in the Long Range Arena (LRA) benchmark~\citep{tay2020long}.

\subsection{Positional Encoding of Transformers}

The attention mechanism of Transformers is defined as:
$$
\textrm{Attn}\left( \mathbf{X}, \mathbf{C} \right) = \sigma\left( \frac{Q(\mathbf{X}) K(\mathbf{C})^\intercal}{\sqrt{d}} \right) V(\mathbf{C})
$$
where $\mathbf{X} \in \mathbb{R}^{l \times d}$ is the query sequence with length $l$, $\mathbf{C} \in \mathbb{R}^{m \times d}$ is the context sequence with length $m$, $\sigma(\cdot)$ is a softmax activation and $Q$, $K$, $V : \mathbb{R}^{d} \rightarrow \mathbb{R}^{d}$ is a linear transformation function projecting inputs into the space of query, key and value respectively. 
Since the attention function is ignorant of the position information of sequence, the Transformer model uses a method that added a special embedding to token embeddings on input of the first layer, called Positional Embedding, to inject position information. \citet{vaswani2017attention} proposed Sinusoidal Positional Embedding, which is a non-trainable constant embedding computed from trigonometric functions. 

On the other hand, BERT \citep{devlin2018bert} uses trainable positional embeddings instead of constant embeddings. It is adaptable to training data, but has limitations such as being unable to handle longer inputs than those used for training and not being translation-invariant.

Relative position methods have been studied, for solving these problems~\citep{shaw2018self,raffel2020exploring}. It learns parameters representing the relative distance between tokens and utilizes them to calculate the attention score. However, It is slower than the sinusoidal approach and uses extra memory and parameters \citep{press2021train}.

\citet{press2021train} pointed out that previous methods are vulnerable to extrapolation and proposes ALiBi, a modified attention function for self-attention as follows:
$$
\textrm{ALiBi}\left( \mathbf{X} \right) = \sigma\left( \frac{Q(\mathbf{X}) K(\mathbf{X})^\intercal}{\sqrt{d}} - \mathbf{D}^\intercal\right) V(\mathbf{X})
$$
$$
\mathbf{D}_{i,j} = \left\{
\begin{array}{ll} 
  m \times (i - j), & \textrm{for } i \ge j \\
  \infty, & \textrm{for } i < j \\
\end{array} 
\right. 
$$
where $m $ is a head-specific positive real-number hyperparameter and $\mathbf{D} \in \mathbb{R}^{l \times l}$ is a distance matrix.

\subsection{Pretraining objectives for Question Answering}
To pretrain a language model, an appropriate training objective that fully exploits the language understanding should be defined. Masked LM \citep{devlin2018bert}, for example, replaces 15\% of the input tokens with a mask token or a random token and forces the model to denoise it. After pre-training is complete, the last hidden representations of the model contains information to restore the replaced token to the original one and it is useful to transfer this information for other NLP tasks as well, such as question answering.

However, Masked LM (MLM) is suboptimal for extractive QA task. \citet{joshi2020spanbert} proposed SpanBERT, which is pretrained by a span-level masking scheme whose lengths follows geometric distribution and it outperformed BERT with MLM in the most of tasks, especially extractive QA. They proved that training objective predicting spans rather than tokens generates better representations especially for span selection tasks.

\citet{ram2021few} introduced Recurring Span Selection (RSS), a novel pre-training objective which is better aligned to QA tasks. In RSS, each recurring text span, except for one to be used as the golden answer span, is masked with a special token, [QUESTION], and a model is trained to point to the position of the golden answer span using the representations from each [QUESTION] token. Because this pre-training task is so similar to the real QA task, the model trained in this objective outperforms models with other pre-training objectives in both the few-shot and high-resource settings for QA.

\subsection{Datasets of Question Answering for Longer Documents}
The most widely used English QA dataset is SQuAD \citep{rajpurkar2016squad}, but it's insufficient to test understanding of long contexts because of its short paragraph. Thus, for QA of longer documents, other datasets are considered. Typical examples are Natural Questions \citep{kwiatkowski2019natural} and TriviaQA \citep{joshi2017triviaqa}, which provide a whole Wikipedia page as the document. NarrativeQA \citep{kovcisky2018narrativeqa}, whose documents consist of movie scripts and books, is another example. Recently, \citet{bowman2022quality} introduced QuALITY, a multiple-choice QA dataset comprised of around 5000 tokens of documents gathered from various sources such as Project Gutenberg and Open American National Corpus.

For Korean QA datasets, the most standard is KorQuAD 1.0 and KorQuAD 2.0, which is comparable to SQuAD in English. The construction and characteristics of the dataset in KorQuAD 1.0 \citep{lim2019korquad1} are nearly identical to those of SQuAD, except that it is in Korean. Therefore, like SQuAD, KorQuAD 1.0 is not suitable for evaluating QA for long documents. To evaluate understanding of longer documents, KorQuAD 2.0 \citep{kim2019korquad} is often used. Since it provides the whole Wikipedia page as a single context without trimming and the page includes not only text but also HTML components such as tables and lists, structural understanding of long HTML documents is required to conquer it. 

\section{LittleBird Architecture}
In this section, we describe the architecture of LittleBird model. Basically, the model can be viewed as a composition of several key ideas including sliding window attention from BigBird \citep{zaheer2020big}, linear bias to attention from ALiBi \citep{press2021train} and pack and unpack attention from LUNA \citep{ma2021luna}.

\subsection{Bidirectional ALiBi}
Since pre-trained language models (PLM) perform best when using data of the same length as the data used for pretraining in general, a new PLM suitable for the length must be built to perform inference on longer data, which is inefficient. To avoid this, we consider the main idea of ALiBi \citep{press2021train}, which is more efficient than relative positional encoding used at T5. However, because ALiBi was designed for causal language modeling, not autoencoding language modeling, each query can attend to keys to the left of itself only, not keys further away or to the right in ALiBi.

Therefore, we devised BiALiBi (Bidirectional ALiBi), which is improved version of ALiBi to suit the autoencoding language model. BiALiBi has the same attention function as ALiBi, but differs only in the method of calculating the distance matrix as follows:
$$
\mathbf{D}_{i,j} = \left\{
\begin{array}{ll} 
  0, & \textrm{for } i = j \\
  \alpha & \textrm{for } i = 0 \textrm{ or } j = 0 \\
  \beta(i - j), & \textrm{for } i > j \\
  \gamma(j - i), & \textrm{for } i < j \\
\end{array} 
\right. 
$$
where $\alpha$, $\beta$ and $\gamma$ are head-specific slopes like $m$ in ALiBi. $\alpha$ is a value for the [CLS] token, which usually appears at position 0. Because this token should be global, it has the same bias regardless of distance. $\beta$ and $\gamma$ are involved in the attention intensity for tokens on the left and right, respectively. Unlike ALiBi, we set $\alpha$, $\beta$ and $\gamma$ as learnable parameters to have more flexibility.

\subsection{Sliding Window Attention}

\begin{figure}
\centering
\includegraphics[width=.5\textwidth]{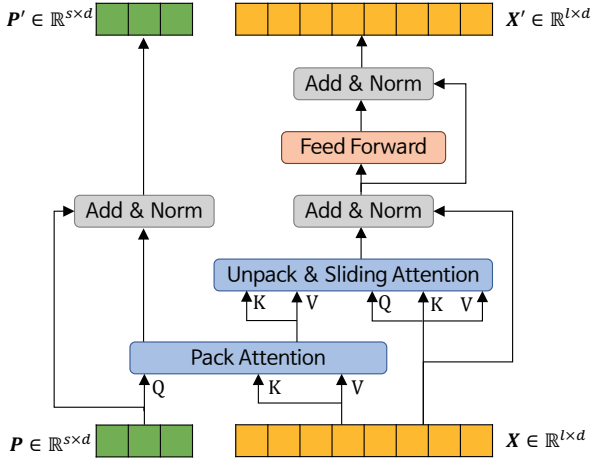}
\caption{\label{fig:lblayer}
LittleBird Layer
}
\end{figure}

\begin{figure}
\centering
\includegraphics[width=.5\textwidth]{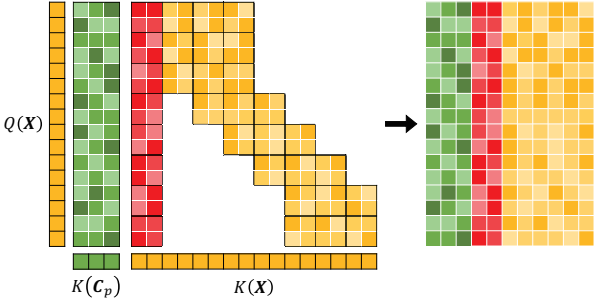}
\caption{\label{fig:lbattention}
Unpack \& Sliding Window Attention of LittleBird
}
\end{figure}

Attention module of BigBird \citep{zaheer2020big} consists of three types of attentions: Global, Window and Random. Global tokens can attend to all other tokens and also can be attended from all other tokens. On the other hand, non-global tokens can only attend to all global tokens, some nearby tokens (Window) and random tokens (Random).

For efficient computation, this attention module is implemented using blocked sliding window. But there is still an overhead where random attention needs repeating gather and scatter operations at random positions. Since it is known that full attention can be substituted well without random attention when global tokens are sufficient \citep{ainslie2020etc}, we completely eliminated random attention from our model. We also reduced the number of global tokens and removed global-local attention, They were replaced with pack and unpack attention, as explained in the following subsection.

\subsection{Pack \& Unpack Attention}

To effectively replace random and global attention, we employed pack and unpack attention \citep{ma2021luna}. However, in the original pack and unpack attention, information loss is unavoidable because all sequence information is packed into a small capacity. We propose adding the sliding window attention to the unpacking step to improve this. Figure \ref{fig:lblayer} depicts the entire architecture of the LittleBird layer.

$$
\mathbf{C}_P = \textrm{Attn}\left(\mathbf{P}, \mathbf{X}\right)
$$
$$
\mathbf{P}' = \textrm{LayerNorm}\left(\mathbf{C}_P + \mathbf{P}\right)
$$
$$
\mathbf{C}_X = \textrm{USWAttn}\left(\mathbf{X}, \mathbf{C}_P \right)
$$
$$
\mathbf{A} = \textrm{LayerNorm}\left(\mathbf{C}_X + \mathbf{X} \right)
$$
$$
\mathbf{X}' = \textrm{LayerNorm}\left(\textrm{FFN}(\mathbf{A}) + \mathbf{A}\right)
$$
\begin{multline*}
\textrm{USWAttn}\left(\mathbf{X}, \mathbf{C}_P \right) = \\
\sigma\left( \frac{Q(\mathbf{X}) \left[K(\mathbf{C}_P) ; K(\mathbf{X})\right]^\intercal}{\sqrt{d}}
- \left[\mathbf{D}_P ; \mathbf{D}\right]^\intercal\right) \\ \cdot \left[V(\mathbf{C}_P) ;V(\mathbf{X})\right]
\end{multline*}
$$
\mathbf{D}_P = \left(\frac{\beta + \gamma}{2}b\right) \mathbf{J}_{s,l}
$$
where $\mathbf{X} \in \mathbb{R}^{l \times d}$ is the input sequence with length $l$,  $\mathbf{P} \in \mathbb{R}^{s \times d}$ is an extra sequence for packing contextual information with length $s$, $\left[\mathbf{A}; \mathbf{B}\right]$ denotes concatenation of matrix $\mathbf{A}$ and $\mathbf{B}$ in row axis, $\mathbf{D} \in \mathbb{R}^{l \times l}$ is a distance matrix from BiALiBi, $\mathbf{D}_P \in \mathbb{R}^{s \times l}$ is a distance matrix for packing tokens and $\mathbf{J}_{s,l}$ is an all-ones matrix with shape $(s, l)$. 

The overall structure is the same as pack and unpack attention \citep{ma2021luna}, but only one part, $\textrm{USWAttn}$ (Unpack \& Sliding Window Attention), is different. In this step, we split $\mathbf{X}$ into blocks with size $b$ and perform block-level attention like \citet{zaheer2020big}, which is demonstrated at Figure \ref{fig:lbattention}. We set only the first block as a global token, and allow local-to-global attention. This is because in most QA tasks, [CLS] tokens and questions are placed in the front part of the input sequence, and we believe it is important to allow the rest of the input sequence to access information of these tokens directly. 

Also, we apply different distance matrices depending on the type of attention. BiALiBi's distance matrix $\mathbf{D}$ is applied to $\mathbf{X}$-to-$\mathbf{X}$ attention, but uniform distance matrix $\mathbf{D}_P$ is applied to $\mathbf{X}$-to-$\mathbf{C}_P$ attention. Since $\mathbf{D}_P$ is defined as a value obtained by multiplying the average of $\beta$ and $\gamma$ by block size $b$, for input tokens, each packing token is considered to be separated by a distance $b$.

Finally, each block can attend itself ($b$ tokens), the blocks next to it ($2b$ tokens), the global tokens($b$ tokens) and the packed context ($s$ tokens). This can be effectively converted to batch matrix multiplication with $O\left(l(4b + s)\right)$ time complexity by rolling and concatenating the key blocks as BigBird. Also, pack attention can be done in $O\left(ls\right)$, so LittleBird attention is done in $O\left(l(4b + 2s)\right)$. If $s$ and $b$ are sufficiently smaller than $l$, it can be considered to have linear time complexity to input length $l$. We choose $s = b = 64$.

\subsection{Efficient Training and Padding Insertion}
\label{sec:training_method}
%
Even with better model structure, pre-training on long inputs is costly. This section describes how to train LittleBird efficiently by reusing an existing PLM. We trained our model by the following steps.
\begin{enumerate}
    \item Initialize all parameters of new LittleBird layers as the corresponding parameters of PLM's layer. Both QKV of pack and of unpack attention are initialized from the attention layer.
    \item Distil knowledge from PLM to the new model using not only soft target probability but also self-attention distribution of each layer as distillation loss following \citet{sun2020mobilebert}, but not hidden states of layers.
    Since long inputs cannot be fed into the teacher model, short inputs are used in this step.
    \item Further train the new model on longer inputs without distillation.
\end{enumerate}

In the step 2, it is important to transfer the self-attention distribution. The parameters of BiALiBi largely dominates the overall attention pattern. We discovered that when we train these parameters without distillation, it takes several epochs to converge, but it converges considerably more stably and fast when we train them with distillation.

When we feed data into the LittleBird model in steps 2 and 3, we can use Padding Insertion (PI), a simple trick to fool the model into thinking it is receiving longer data even though receiving shorter data actually. Consider the following scenario: a virtual padding token is randomly inserted in the middle of input data. Padding tokens are masked because they should not be attended by other tokens. As a result, inserting padding has no effect on other tokens, but only on the computing distance matrix of BiALiBi. So, instead of inserting padding tokens actually, we can achieve the same result by manipulating the position ids of input sequence for the distance matrix. This allows us to train the model to prepare for long inputs while actually taking short inputs. It makes the extrapolation capability of the model robust, and the experimental results for this are given in Section \ref{sec:padding_insertion}.

Also, it may be important to balance the epochs of steps 2 and 3 for efficient training. We obtained good results in the following experiments by simply allocating the same epochs to steps 2 and 3, but better balancing is worth further study.

\section{Experiment}
The goal of this section is to demonstrate the benefits of LittleBird compared to other long transformer models. First we check if the proposed training method works well with the architecture of LittleBird. Following that, we consider QA tasks of English and Korean, which requires longer sequence modeling. Lastly we prove that Padding Insertion significantly improves accuracy and that the efficiency of our model by performing training and inference for various input lengths.

\subsection{Pre-training using RSS}
\label{sec:pretraining_rss}

\begin{table*}[]
\centering
\begin{tabular}{lccc}
\hline
\textbf{Model} & \textbf{512} & \textbf{2k} & \textbf{4k}  \\
\hline
\textbf{A}: LittleBird, distilled with 512 length & 84.46 & 42.02 & 32.08 \\
\textbf{B}: LittleBird, trained with 2k length from \textbf{A} & 73.68 & 72.67 & 47.92 \\
\textbf{C}: LittleBird, trained with 4k length from \textbf{A} & 65.34 & 64.08 & 63.59 \\ 
\hline
\textbf{D}: BigBird, trained with 4k length & 66.04 & 48.97 & 58.30 \\
\hline
\end{tabular}
\caption{\label{tab:pretraining_acc} RSS pre-training accuracy of English dev datasets. Each column represents the sequence length of the dev datasets.}
\end{table*}
We pre-trained the model with the RSS objective \citep{ram2021few} to create a English and a Korean model optimized for QA. For pre-training the English model, RoBERTa checkpoint\footnote{\url{https://huggingface.co/roberta-base}} was used for initializing and it was also used as the teacher model in the procedure described in Section \ref{sec:training_method}. Three public datasets, OpenWebText\citep{Gokaslan2019OpenWeb}, CC-News\citep{guu2020retrieval} and English Wikipedia were used. Similarly, KoELECTRA checkpoint \footnote{\url{https://github.com/monologg/KoELECTRA}} was used for the Korean model, and various Korean corpora including Wikipedia were used. Details for pre-training are attached in the Appendix \ref{sec:ap_experiment}.

Table \ref{tab:pretraining_acc} displays the accuracy of the four English pre-trained models on dev datasets of 512, 2k and 4k lengths. An accuracy of RSS tasks is defined as the proportion in which the span selected by a model exactly matches the golden answer. The four models \textbf{A}, \textbf{B}, \textbf{C} and \textbf{D} were pre-trained using different training settings. 
The LittleBird model \textbf{A} was trained using only steps 1 and 2 in Section \ref{sec:training_method}, and the models \textbf{B} and \textbf{C} were further trained from \textbf{A} using step 3. The model \textbf{D} was trained using warm-starting from the same RoBERTa checkpoint without any distillations in BigBird architecture for the same amount of time as \textbf{B} and \textbf{C}.
All results of the table are without Padding Insertion. It is noteworthy that the LittleBird model \textbf{C} achieved higher pre-training accuracy than the BigBird model \textbf{D} at 2k and 4k dev datasets when trained at the same time, and \textbf{B} and \textbf{C} loses their accuracy for short inputs as trained for longer inputs.

\subsection{QA Benchmark for English}
\label{sec:qa_for_english}

\begin{table*}[]
\centering
\begin{tabular}{cccccccc}
\hline
\multirow{2}{*}{\textbf{Model}} & \multicolumn{3}{c}{\textbf{HotpotQA}} & \multicolumn{2}{c}{\textbf{NaturalQ}}  & \multicolumn{1}{c}{\textbf{TriviaQA}} & \multicolumn{1}{c}{\textbf{Wikihop}} \\
 & \multicolumn{1}{c}{\textbf{Ans}} & \multicolumn{1}{c}{\textbf{Sup}} & \multicolumn{1}{c}{\textbf{Joint}} & \multicolumn{1}{c}{\textbf{LA}} & \multicolumn{1}{c}{\textbf{SA}} & \multicolumn{1}{c}{\textbf{Full}} & \multicolumn{1}{c}{\textbf{MCQ}} \\
\hline
Longformer & 74.3 & 84.4 & 64.4 & - & - & 75.2 & 75.0 \\
BigBird & 75.7 & 86.8 & 67.7 & 70.8 & 53.3 & \textbf{79.5} & 75.9 \\
ETC & 75.5 & \textbf{87.1} & 67.8 & 73.9 & 54.9 & 78.7 & 75.9 \\
\hline
BigBird + RSS & 73.7 & 83.2 & 63.2 & 71.1 & 51.2 & 75.5 & 75.1 \\
LittleBird & \textbf{77.7} & 86.3 & \textbf{68.5} & \textbf{76.7} & \textbf{57.5} & 76.9 & \textbf{82.0} \\
\hline
\end{tabular}
\caption{\label{tab:english_qas} QA Dev results using base-size models. We report accuracy for WikiHop and F1 for the others.
}
\end{table*}

\begin{table*}[]
\centering
\begin{tabular}{cccccccc}
\hline
\multirow{2}{*}{\textbf{Pack Size}} & \multicolumn{3}{c}{\textbf{HotpotQA}} & \multicolumn{2}{c}{\textbf{NaturalQ}}  & \multicolumn{1}{c}{\textbf{TriviaQA}} & \multicolumn{1}{c}{\textbf{Wikihop}} \\
 & \multicolumn{1}{c}{\textbf{Ans}} & \multicolumn{1}{c}{\textbf{Sup}} & \multicolumn{1}{c}{\textbf{Joint}} & \multicolumn{1}{c}{\textbf{LA}} & \multicolumn{1}{c}{\textbf{SA}} & \multicolumn{1}{c}{\textbf{Full}} & \multicolumn{1}{c}{\textbf{MCQ}} \\
\hline
0 & 69.0 & 77.0 & 54.6 & 73.3 & 54.1 & 71.8 & 73.9 \\
32 & 72.2 & 83.0 & 61.7 & 73.4 & 55.1 & 72.5 & 77.1 \\
64 & \textbf{77.7} & \textbf{86.3} & \textbf{68.5} & \textbf{76.7} & \textbf{57.5} & \textbf{76.9} & \textbf{82.0} \\
\hline
\end{tabular}
\caption{\label{tab:english_qas_pack} QA Dev results of LittleBird with varying pack size. We report accuracy for WikiHop and F1 for the others.
}
\end{table*}

In this section, we compare the performance of LittleBird with other models for long inputs, such as Longformer \citep{beltagy2020longformer}, BigBird and ETC. We used the LittleBird models pre-trained on 2048-length with Padding Insertion (PI) in this experiments. The English BigBird model was also trained in the same setting as LittleBird using RSS for comparison. Experiments were performed on four QA datasets, HotpotQA \citep{yang2018hotpotqa}, NaturalQ\cite{kwiatkowski2019natural}, TriviaQA\cite{joshi2017triviaqa} and WikiHop\citep{welbl2018constructing}, used by \citet{zaheer2020big}. Because the most of datasets have long input data of more than 4K tokens, they are appropriate for evaluating long text understanding. For detailed statistics on these datasets, see Table \ref{tab:dataset_stat} in the appendix.
The top three rows of the Table \ref{tab:english_qas} are from previous papers' results, and the row \textbf{BigBird + RSS} denotes the model pre-trained using RSS objective with the same training time as LittleBird in Section \ref{sec:pretraining_rss}. LittleBird shows more effectiveness when computational resources and time are limited and achieved higher accuracy than other models in the most of cases.

The same experiment was repeated while changing the pack size $s$ to examine the effect of pack and unpack attention, and the results are shown in Table \ref{tab:english_qas_pack}. When $s=0$, LittleBird model is equivalent to a model with only sliding window attention and local-global attention. It can be confirmed that pack and unpack attention is significantly effective in all cases.

\subsection{QA Benchmark for Korean}

\begin{table}[]
\centering
\begin{tabular}{lccc}
\hline
\textbf{Model / Seq. Len.} & \textbf{EM} & \textbf{F1} \\
\hline
\makecell[l]{KLUE-RoBERTa / 512 \\ \citep{park2021klue}}  & 55.44 & 73.02 \\
\makecell[l]{KoBigBird / 4K \\ \citep{jangwon_park_2021_5654154}} & 67.77 & 82.03 \\
\hline
KoELECTRA / 512 \citep{park2020koelectra}  & 66.95 & 77.75 \\
KoELECTRA + RSS / 512 & 68.56 & 79.30 \\
\hline
LittleBird / 512  & 68.11 & 80.38 \\
\quad\quad\quad\quad / 1K   & 72.82 & 83.09 \\
\quad\quad\quad\quad / 2K   & 75.50 & 84.78 \\
\quad\quad\quad\quad / 4K   & 76.01 & 85.20 \\
\quad\quad\quad\quad / 8K   & \textbf{77.79} & \textbf{89.39} \\
\hline
\end{tabular}
\caption{\label{tab:korquad_dev}
KorQuAD2.0 Dev results using Base size models
}
\end{table}

\begin{table}[]
\centering
\begin{tabular}{cccc}
\hline
\textbf{Model} & \textbf{EM} & \textbf{F1} & \textbf{Latency} \\
\hline
SDS-Net v1.3 & 77.86 & 89.92 & 10.43 \\
Ko-LongBERT & 77.88 & 89.62 & 10.05 \\
SkERT-Large 1.1 & 77.44 & 88.81 & 10.05 \\
\hline
LittleBird-Base & 76.66 & 88.57 & 2.29 \\
LittleBird-Large & \textbf{78.70} & \textbf{90.22} & 6.16 \\
\hline
\end{tabular}
\caption{\label{tab:korquad_test}
KorQuAD2.0 Test results with previous SOTAs. Latency is a measure of the average response time per question in seconds.
}
\end{table}


First, we compared the base size LittleBird model with other Korean PLM using dev set of KorQuAD2.0, and the results are shown in Table \ref{tab:korquad_dev}. For KLUE-RoBERTa and KoBigBird, we described the reported performance, however for KoELECTRA, we conducted fine-tuning and evaluation using the published model. In the table, \textbf{KoELECTRA} denotes the results of using the published model as is, and \textbf{KoELECTRA + RSS} denotes the results after further training with RSS.
Also, on the same LittleBird model, we repeated the experiment by varying the maximum sequence length of fine-tuning. The LittleBird model fine-tuned on 8K showed the best performance and this clearly demonstrates the significance of broad context when performing QA on long documents.

A large model was trained using the best setting of the base-size model, we submitted both the base and the large size model to the evaluation system, and the results are shown in Table \ref{tab:korquad_test}\footnote{The whole list is available at \url{https://korquad.github.io/}}. The models listed in the table with LittleBird are the previous top three models. Although these models' exact structure or size is not disclosed, it can be confirmed that LittleBird is faster and more accurate than them.

\subsection{Effect of Padding Insertion}
\label{sec:padding_insertion}
In addition, experiments were performed to determine whether the Padding Insertion (PI) introduced in Section \ref{sec:training_method} is effective. First, we verified that the model fine-tuned with PI performs well with inputs that are longer than those used for fine-tuning. We chose a Korean fake news dataset\citep{lee2019fake} for fine-tuning, which requires binary classification of whether a given text is fake or not. Since the dataset's average token length of 774 and maximum token length of 17,488 and it contains many documents longer than 512 tokens, it was appropriate for evaluating accuracy on long inputs. For this experiment, the Korean LittleBird in base size, pre-trained without PI, was used. Table \ref{tab:pad_inserting_fakenews} displays the result. Because this dataset's label distribution is slightly skewed, we presented the ratio of true labels for each interval in the table so that the accuracy at each interval may be compared more equitably.

\begin{table}[]
\centering
\begin{tabular}{ccccc}
\hline
\textbf{Seq. Len.} & \textbf{512} & \textbf{1024} & \textbf{1536} & \textbf{2048} \\
\hline
512 & 99.53 & 48.17 & 47.31 & 74.23 \\
1K & 99.49 & \textbf{99.15} & 98.21 & 88.66 \\
\hline
512 + PI & 99.36 & 97.67 & 96.06 & 95.88 \\
1K + PI & \textbf{99.57} & 99.11 & \textbf{98.92} & \textbf{96.91} \\
\hline
True labels & 44.7\% & 52.5\% & 56.3\% & 27.6\% \\ 
\hline
\end{tabular}
\caption{\label{tab:pad_inserting_fakenews} Classification accuracy with varying max sequence length of fine-tuning Korean fake news data. Each column represents the sequence length of the dev dataset. The last row represents the distribution of true labels for documents in each interval.
}
\end{table}
The row in the table with \textbf{+ PI} indicates that PI was used in the fine-tuning phase. Paddings with a length between 0 and 256 were inserted randomly with a 20\% probability at the end of each sentence (after the `.', `?' and `!' token).
When the model was fine-tuned on 512-length sequences, it did not work well for inputs longer than 512. This result is consistent to the prior study \citep{press2021train} that found that the extrapolation performance is poor when the head-specific slope parameters of ALiBi are learnable. However, when fine-tuned with PI on 512-length sequences, the accuracy was comparable to one fine-tuned on sequences of 1024-length and the model fine-tuned with PI on 1024-length sequences show the best accuracy. The results show that PI can improve the model's extrapolation performance for longer inputs. 

\begin{table*}[]
\centering
\begin{tabular}{cccccccc}
\hline
\multirow{2}{*}{\textbf{Seq. Len.}} & \multicolumn{3}{c}{\textbf{HotpotQA}} & \multicolumn{2}{c}{\textbf{NaturalQ}}  & \multicolumn{1}{c}{\textbf{TriviaQA}} & \multicolumn{1}{c}{\textbf{Wikihop}} \\
 & \multicolumn{1}{c}{\textbf{Ans}} & \multicolumn{1}{c}{\textbf{Sup}} & \multicolumn{1}{c}{\textbf{Joint}} & \multicolumn{1}{c}{\textbf{LA}} & \multicolumn{1}{c}{\textbf{SA}} & \multicolumn{1}{c}{\textbf{Full}} & \multicolumn{1}{c}{\textbf{MCQ}} \\
\hline
512 & 75.3 & 83.5 & 64.2 & 74.3 & 55.8 & 70.1 & 77.8 \\
2K & 75.5 & 85.3 & 65.9 & 74.8 & 55.2 & 74.6 & 79.7 \\
4K & 74.3 & 84.8 & 64.7 & 74.8 & 55.9 & 74.3 & 79.1 \\
\hline
512 + PI & 76.1 & 85.6 & 66.6 & 74.9 & 57.3 & 75.8 & 79.9 \\
2K + PI & \textbf{77.7} & \textbf{86.3} & \textbf{68.5} & \textbf{76.7} & \textbf{57.5} & \textbf{76.9} & \textbf{82.0} \\
\hline
\end{tabular}
\caption{\label{tab:pad_inserting_english_qa} QA Dev results with varying max sequence length of pre-training data. We report accuracy for WikiHop and F1 for the others.
}
\end{table*}

We also performed experiments about the effect of PI in the pre-training stage. We looked at the four QA datasets used in Section \ref{sec:qa_for_english}.
Table \ref{tab:pad_inserting_english_qa} shows the effect of sequence length and PI in the pre-training stage for the English LittleBird model. Pre-trained models on longer inputs did not always perform well. Models with PI, on the other hand, consistently outperformed those without. Particularly, the model pre-trained on 2048-length with PI outperforms the other models. Pre-training on long inputs does not guarantee accuracy on shorter inputs, as shown in Table \ref{tab:pretraining_acc}. When PI is used, all inputs will have varying lengths, causing the model to encounter inputs of varying lengths, which makes the model more robust.

\subsection{Ablation Study}

\begin{table*}[]
\centering
\begin{tabular}{lccccccc}
\hline
\multirow{2}{*}{\textbf{Model}} & \multicolumn{3}{c}{\textbf{HotpotQA}} & \multicolumn{2}{c}{\textbf{NaturalQ}}  & \multicolumn{1}{c}{\textbf{TriviaQA}} & \multicolumn{1}{c}{\textbf{Wikihop}} \\
 & \multicolumn{1}{c}{\textbf{Ans}} & \multicolumn{1}{c}{\textbf{Sup}} & \multicolumn{1}{c}{\textbf{Joint}} & \multicolumn{1}{c}{\textbf{LA}} & \multicolumn{1}{c}{\textbf{SA}} & \multicolumn{1}{c}{\textbf{Full}} & \multicolumn{1}{c}{\textbf{MCQ}} \\
\hline
LittleBird & \textbf{77.7} & \textbf{86.3} & \textbf{68.5} & \textbf{76.7} & \textbf{57.5} & \textbf{76.9} & \textbf{82.0} \\
- BiALiBi & 75.2 & 85.2 & 65.6 & 74.3 & 55.4 & 76.5 & 78.3 \\
- Sliding Window Attn & 16.1 & 27.9 & 6.0 & 11.8 & 1.4 & 33.9 & 20.3 \\
- Pack Attn & 69.0 & 77.0 & 54.6 & 73.3 & 54.1 & 71.8 & 73.9 \\
- Padding Insertion & 75.5 & 85.3 & 65.9 & 74.8 & 55.2 & 74.6 & 79.7 \\
\hline
\end{tabular}
\caption{\label{tab:ablation_study} Ablation study of LittleBird. We report accuracy for WikiHop and F1 for the others.
}
\end{table*}
We conducted an ablation study to assess how much each component contributed to the LittleBird model's performance, and the results are shown in Table \ref{tab:ablation_study}. As can be seen from the table, the factors that have the biggest impact on performance are Pack Attention and Sliding Window Attention. And in the case of BiALiBi, its contribution to performance improvements is small but it helps to replace Absolute Positional Embedding, allowing it to accept variable-length inputs. It can be seen that applying Padding Insertion without changing the structure of the model makes an additional improvement by itself.

\subsection{Speed \& Memory efficiency}

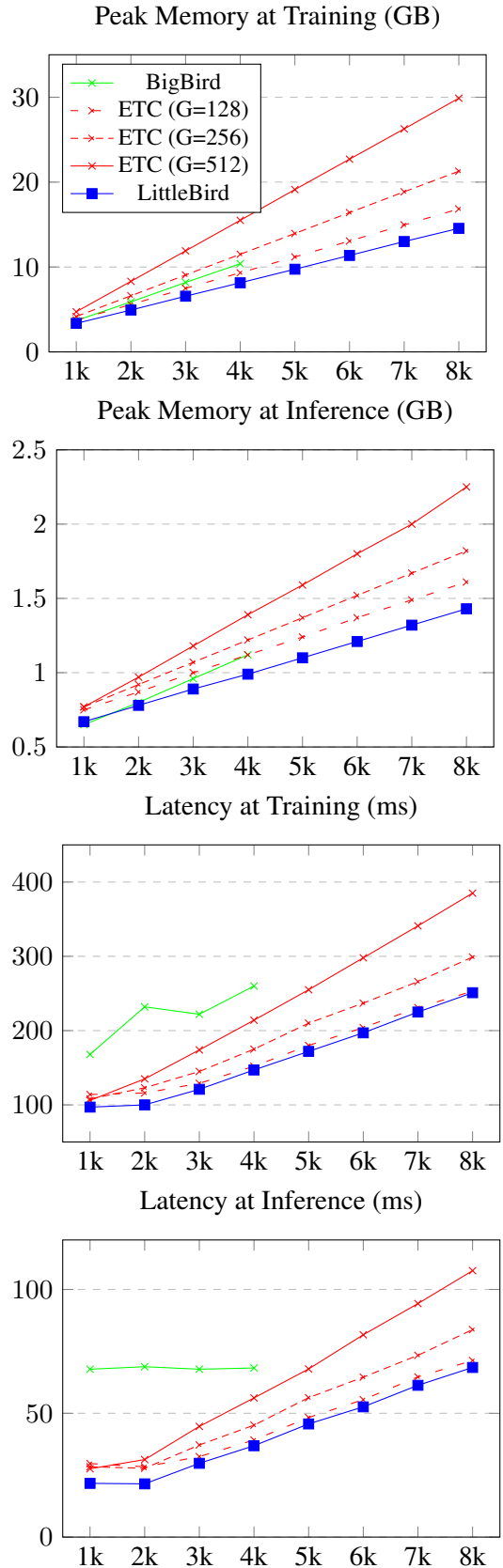
\begin{figure}
    \begin{subfigure}[b]{1\columnwidth}
    \begin{tikzpicture}
    \begin{axis}[
        title={Peak Memory at Training (GB)},
        width={1\columnwidth},
        height={0.75\columnwidth},
        legend style={nodes={scale=0.85, transform shape}},
        xmin=0.5, xmax=8.5,
        ymin=0, ymax=35,
        xtick={1,2,3,4,5,6,7,8},
        xticklabels={1k,2k,3k,4k,5k,6k,7k,8k},
        legend pos=north west,
        ymajorgrids=true,
        grid style=dashed,
    ]
    
    \addplot[color=green, mark=x,]
        coordinates {
        (1, 3.69)(2, 5.89)(3, 8.18)(4, 10.39)
        };
        \addlegendentry{BigBird}
    
    \addplot[color=red, mark=x, loosely dashed]
        coordinates {
        (1, 3.75)(2, 5.58)(3, 7.49)(4, 9.32)(5, 11.19)(6, 13.06)(7, 14.96)(8, 16.84)
        };
        \addlegendentry{ETC (G=128)}
    
    \addplot[color=red, mark=x, densely dashed]
        coordinates {
        (1, 4.17)(2, 6.61)(3, 9.06)(4, 11.50)(5, 13.95)(6, 16.41)(7, 18.86)(8, 21.28)
        };
        \addlegendentry{ETC (G=256)}
    
    \addplot[color=red, mark=x,]
        coordinates {
        (1, 4.73)(2, 8.31)(3, 11.90)(4, 15.50)(5, 19.12)(6, 22.70)(7, 26.27)(8, 29.88)
        };
        \addlegendentry{ETC (G=512)}
    
    \addplot[color=blue, mark=square*]
        coordinates {
        (1, 3.36)(2, 4.93)(3, 6.56)(4, 8.14)(5, 9.74)(6, 11.36)(7, 13.00)(8, 14.55)
        };
        \addlegendentry{LittleBird}
    
    \end{axis}
    \end{tikzpicture}
    \end{subfigure}
    \hfill
    \begin{subfigure}[b]{1\columnwidth}
    \begin{tikzpicture}
    \begin{axis}[
        title={Peak Memory at Inference (GB)},
        width={1\columnwidth},
        height={0.75\columnwidth},
        xmin=0.5, xmax=8.5,
        ymin=0.5, ymax=2.5,
        xtick={1,2,3,4,5,6,7,8},
        xticklabels={1k,2k,3k,4k,5k,6k,7k,8k},
        legend pos=north west,
        ymajorgrids=true,
        grid style=dashed,
    ]
    
    \addplot[color=green, mark=x,]
        coordinates {
        (1, 0.65)(2, 0.80)(3, 0.96)(4, 1.12)
        };
        \addlegendentry{BigBird}
    
    \addplot[color=red, mark=x, loosely dashed]
        coordinates {
        (1, 0.75)(2, 0.87)(3, 1.00)(4, 1.12)(5, 1.24)(6, 1.37)(7, 1.49)(8, 1.61)
        };
        \addlegendentry{ETC (G=128)}
    
    \addplot[color=red, mark=x, densely dashed]
        coordinates {
        (1, 0.77)(2, 0.92)(3, 1.07)(4, 1.22)(5, 1.37)(6, 1.52)(7, 1.67)(8, 1.82)
        };
        \addlegendentry{ETC (G=256)}
    
    \addplot[color=red, mark=x,]
        coordinates {
        (1, 0.77)(2, 0.97)(3, 1.18)(4, 1.39)(5, 1.59)(6, 1.80)(7, 2.00)(8, 2.25)
        };
        \addlegendentry{ETC (G=512)}
    
    \addplot[color=blue, mark=square*]
        coordinates {
        (1, 0.67)(2, 0.78)(3, 0.89)(4, 0.99)(5, 1.10)(6, 1.21)(7, 1.32)(8, 1.43)
        };
        \addlegendentry{LittleBird}
    \legend{}
    \end{axis}
    \end{tikzpicture}
    \end{subfigure}
    \hfill
    \begin{subfigure}[b]{1\columnwidth}
    \begin{tikzpicture}
    \begin{axis}[
        title={Latency at Training (ms)},
        width={1\columnwidth},
        height={0.75\columnwidth},
        xmin=0.5, xmax=8.5,
        ymin=50, ymax=450,
        xtick={1,2,3,4,5,6,7,8},
        xticklabels={1k,2k,3k,4k,5k,6k,7k,8k},
        legend pos=north west,
        ymajorgrids=true,
        grid style=dashed,
    ]
    
    \addplot[color=green, mark=x,]
        coordinates {
        (1, 168)(2, 232)(3, 222)(4, 260)
        };
        \addlegendentry{BigBird}
    
    \addplot[color=red, mark=x, loosely dashed]
        coordinates {
        (1, 114)(2, 116)(3, 129)(4, 152)(5, 180)(6, 204)(7, 231)(8, 253)
        };
        \addlegendentry{ETC (G=128)}
    
    \addplot[color=red, mark=x, densely dashed]
        coordinates {
        (1, 108)(2, 123)(3, 145)(4, 175)(5, 210)(6, 237)(7, 266)(8, 299)
        };
        \addlegendentry{ETC (G=256)}
    
    \addplot[color=red, mark=x,]
        coordinates {
        (1, 106)(2, 135)(3, 174)(4, 214)(5, 255)(6, 298)(7, 341)(8, 385)
        };
        \addlegendentry{ETC (G=512)}
    
    \addplot[color=blue, mark=square*]
        coordinates {
        (1, 97)(2, 100)(3, 121)(4, 147)(5, 172)(6, 197)(7, 225)(8, 251)
        };
        \addlegendentry{LittleBird}
    \legend{}
    \end{axis}
    \end{tikzpicture}
    \end{subfigure}
    \begin{subfigure}[b]{1\columnwidth}
    \begin{tikzpicture}
    \begin{axis}[
        title={Latency at Inference (ms)},
        width={1\columnwidth},
        height={0.75\columnwidth},
        xmin=0.5, xmax=8.5,
        ymin=0, ymax=120,
        xtick={1,2,3,4,5,6,7,8},
        xticklabels={1k,2k,3k,4k,5k,6k,7k,8k},
        legend pos=south east,
        ymajorgrids=true,
        grid style=dashed,
    ]
    
    \addplot[color=green, mark=x,]
        coordinates {
        (1, 67.8)(2, 68.8)(3, 67.8)(4, 68.3)
        };
        \addlegendentry{BigBird}
    
    \addplot[color=red, mark=x, loosely dashed]
        coordinates {
        (1, 29.7)(2, 28.4)(3, 32.5)(4, 39.2)(5, 48.3)(6, 55.5)(7, 64.7)(8, 71.3)
        };
        \addlegendentry{ETC (G=128)}
    
    \addplot[color=red, mark=x, densely dashed]
        coordinates {
        (1, 28.4)(2, 27.8)(3, 37.2)(4, 45.2)(5, 56.3)(6, 64.6)(7, 73.4)(8, 83.8)
        };
        \addlegendentry{ETC (G=256)}
    
    \addplot[color=red, mark=x,]
        coordinates {
        (1, 27.6)(2, 31.3)(3, 44.7)(4, 56.2)(5, 67.9)(6, 81.7)(7, 94.3)(8, 107.6)
        };
        \addlegendentry{ETC (G=512)}
    
    \addplot[color=blue, mark=square*]
        coordinates {
        (1, 21.7)(2, 21.5)(3, 29.8)(4, 36.9)(5, 45.7)(6, 52.6)(7, 61.3)(8, 68.5)
        };
        \addlegendentry{LittleBird}
    \legend{}
    \end{axis}
    \end{tikzpicture}
    \end{subfigure}
\caption{\label{fig:peakmem_latency}
Peak memory usage and latency at training and inference in a single batch with varying input sequence lengths. Measured at PyTorch 1.8.1 \& CUDA 11.1 environment on a single NVIDIA A100.
(G=$n$) indicates the number of global tokens in ETC model.
}
\end{figure}

To confirm the efficiency of LittleBird empirically, we investigate the speed and memory footprint of base-size BigBird, ETC and LittleBird with varying input lengths (1K to 4K for BigBird, 1K to 8K for others). All models are evaluated on simple masked-language model task with the same single batch. The result is shown in Figure \ref{fig:peakmem_latency}.

In all three models, memory and computation time increase in proportion to input length, and in the case of the ETC model, the number of global tokens has a significant impact on their overall efficiency. \footnote{In the case of model BigBird, as an exception, the inference latency did not increase in proportion to input length. Since the same results were obtained in several repeated experiments, it is assumed that there is an overhead in the BigBird implementation of transformers 4.6.1, which is used in the experiment.} It is noteworthy that the authors set the number of global tokens in the range of 230 to 430 when testing on QA datasets, it is clear that the result between $G=256$ and $G=512$ is the performance of a practically usable ETC model. In all cases, the efficiency of LittleBird far outperforms that of the other two models.

\section{Conclusion}
We propose LittleBird, which is more efficient in terms of memory and computational time than existing Transformer models for long sequences, and its effective way to train. It combines a novel position encoding method, BiALiBi, and pack \& unpack with sliding window attention to achieve high speed and accuracy, particularly in question answering tasks for long documents. The distillation and training method with Padding Insertion allows the model to be trained by reusing the existing pre-trained language model for short inputs and work well for long inputs even if trained on short inputs. We demonstrated through experiments that the accuracy of question answering improves as the model is fed a longer input, and we achieved state-of-the-art performance in KorQuAD2.0 using LittleBird.

\section*{Limitations}

First, despite the efficiency of LittleBird and its excellent performance in question answering, it is still unknown whether LittleBird works well for other NLP tasks. Further research is needed on other tasks. Second, in the encoder-decoder architecture model that requires cross-attention, since position information is not injected into each token at all, it may be difficult for the decoder layer to find appropriate tokens of the encoder to attend. Lastly, causal masking cannot be applied to the pack and unpack attention due to its characteristics, which means that LittleBird cannot be used to the decoder-only autoregressive language model.

\section*{Ethics Statement}
No private data is used at all, and all the datasets used in this paper for pre-training and fine-tuning are accessible to the public. Thus, our work is free of any privacy issues.


%

\bibliography{ms}
\bibliographystyle{acl_natbib}

\newpage
\appendix


\section{Experiments details}
\label{sec:ap_experiment}

\subsection{Pre-training}
We used three publicly available datasets OpenWebText\citep{Gokaslan2019OpenWeb}, CC-News\citep{guu2020retrieval} and Wikipedia to pre-train English LittleBird model. The RoBERTa model's vocabulary was borrowed, and the weights were also initialized from the same RoBERTa checkpoint\footnote{\url{https://huggingface.co/roberta-base}}. 
For Korean LittleBird model, Newspaper, Written, Web corpus from 모두의 말뭉치(Moduui Malmungchi)\footnote{\url{https://corpus.korean.go.kr/}} and Wikipedia were used, we borrowed the vocabulary of KoELECTRA\citep{park2020koelectra} and used its weights to initialize.

Following \citet{ram2021few}, we preprocessed the corpora to find recurring spans. In the cluster selection step, up to 30 spans were dynamically selected for each 512-token document, 120 spans for each 2K-token document and 240 spans for each 4K-token document. QASS was used as a pre-training architecture, but instead of their proposal of predicting the start and the end positions of the answer independently, we predict the start position first and then jointly predict the end position conditionally on the start position.

\subsection{Fine-tuning on Question Answering tasks}
\begin{table*}[]
\centering
\begin{tabular}{cccccc}
\hline
\textbf{Lang.} & \textbf{Dataset} & \multicolumn{2}{c}{\textbf{\# of Examples}} & \multicolumn{2}{c}{\textbf{Sequence Length in Tokens}} \\
 & & \textbf{Train} & \textbf{Dev} & \textbf{0 / 25 / 50 / 75 / 100th Perc.} & \textbf{Avg.} \\
\hline
English & \makecell[c]{HotpotQA-distractor \\ \citep{yang2018hotpotqa}} & 90447 & 7405 & 59 / 1063 / 1259 / 1476 / 3717 & 1284 \\
& \makecell[c]{Natural Questions \\ \citep{kwiatkowski2019natural}} & 307373 & 7830 & 269 / 3852 / 7371 / 14120 / 147467 & 10485 \\
& \makecell[c]{TriviaQA \\ \citep{joshi2017triviaqa}} & 61888 & 7993 & 32 / 3687 / 8828 / 16469 / 177156 & 11789 \\
& \makecell[c]{WikiHop \\ \citep{welbl2018constructing}} & 43738 & 5129 & 78 / 759 / 1308 / 2076 / 19802 & 1563 \\
\hline
Korean & \makecell[c]{KorQuAD2.0 \\ \citep{kim2019korquad}} & 83486 & 10165 & 194 / 1661 / 2813 / 4985 / 20557 & 4411 \\
\hline
\end{tabular}
\caption{\label{tab:dataset_stat} Statistics of Question Answering datasets}
\end{table*}

We basically applied QASS on the model architectures for question answering used by \citet{zaheer2020big}. The task-specific detailed structure is as follows.

\textbf{HotpotQA}: We place the [CLS] token, the [QUESTION] token and the question in sequence, and then each paragraph separated by [SEP] is followed. Also, 32 virtual paddings are inserted between paragraphs for LittleBird model. A linear layer is used to predict the start position of the answer, and a concatenation of the representations of the start position and of the end position is fed to double layer consisting of a gelu-activated layer and a linear layer to predict the ending position. For evidence classification, a concatenation of the representations of the start and end 
positions of each evidence is fed into a double layer with a gelu-activation. Finally, for answer type classification, the representation of the [CLS] token is fed into a double layer with a gelu-activation.

\textbf{NaturalQ}: We place the [CLS] token, the [QUESTION] token and the question like HotpotQA, and the whole paragraph is followed. A sliding window with 4K-length stride was used for paragraphs exceeding 8K in length. For short answers, a model predicts the start position first, and then predicts the end position by concatenating the representation of start and the end position, similarly to HotpotQA. To predict long answers, a concatenation of the representation of the start and the end position is fed to double layer with a gelu-activation. Also, the representation of the [CLS] token is used for answer type classification like HotpotQA.

\textbf{TriviaQA}: We place the [CLS] token, the [QUESTION] token and the question, and then each paragraph separated by [SEP] is followed. Also, 32 virtual paddings are inserted between paragraphs for LittleBird model. A sliding window with 4K-length stride was used for paragraphs exceeding 8K in length. For training noisy spans we follow \citet{clark2017simple}. To predict the start and the end positions of answers, we use the same predictor as HotpotQA's one.

\textbf{WikiHop}: We place the [CLS] token, the [QUESTION] token, and the question, and then each answer separated by [SEP] is followed next, and each paragraph separated by [SEP] is followed lastly. Also, 32 virtual paddings are inserted between paragraphs for LittleBird model. To predict the start and the end positions of answers, we use the same predictor as HotpotQA's one.

\textbf{KorQuAD2.0}: We place the [CLS] token, the [QUESTION] token and the question like HotpotQA, and the whole paragraph is followed. A sliding window with 4K-length stride was used for paragraphs exceeding 8K in length. Also we add tokens (<H1>, <H2>, <P>, <Table>, <Tr>, <Td>, <Th>, <Ul>, <Ol>, <Li>) for key HTML elements to the pre-trained model's vocabulary. Also, HTML documents of KorQuAD2 was preprocessd to remove unnecessary headers, footers, script, and style tags. To predict the start and the end positions of answers, we use the same predictor as HotpotQA's one.

Table \ref{tab:dataset_stat} shows the statistics of QA dataset used in this paper. Sequence lengths were measured in sub-word tokens and tabulated with averages and 0 to 100 percentiles. Table \ref{tab:english_qa_hp} displays the hyperparameters for English LittleBird that were used to create Tables \ref{tab:english_qas} and \ref{tab:english_qas_pack}. Hyperparameters for Korean Littlebird, used for creating Table \ref{tab:korquad_dev} and \ref{tab:korquad_test} are shown in Table \ref{tab:korquad_hp}.

\begin{table*}[]
\centering
\begin{tabular}{ccccc}
\hline
\textbf{Parameter} & \textbf{HotpotQA} & \textbf{NaturalQ} & \textbf{TriviaQA} & \textbf{WikiHop} \\
\hline
Pack Size & \multicolumn{4}{c}{64} \\
Block Size & \multicolumn{4}{c}{64} \\
Max Seq. Length & 4096 & 8192 & 8192 & 8192 \\
\# of heads & \multicolumn{4}{c}{12} \\
\# of hidden layers & \multicolumn{4}{c}{12} \\
Hidden layer size & \multicolumn{4}{c}{768} \\
Total Parameters & \multicolumn{4}{c}{145M} \\
Batch size & 64 & 128 & 128 & 32 \\
Optimizer & \multicolumn{4}{c}{AdamW} \\
Learning rate & \multicolumn{4}{c}{5e-5} \\
Compute resources & \multicolumn{4}{c}{4 $\times$ NVIDIA A100} \\
\hline
\end{tabular}
\caption{\label{tab:english_qa_hp} Hyperparameters of base LittleBird model used for English Question Answering}
\end{table*}

\begin{table}[]
\centering
\begin{tabular}{ccccc}
\hline
\textbf{Parameter} & \textbf{Base} & \textbf{Large} \\
\hline
Pack Size & \multicolumn{2}{c}{64} \\
Block Size & \multicolumn{2}{c}{64} \\
Max Seq. Length & 6144 & 8192 \\
\# of heads & 12 & 16 \\
\# of hidden layers & 12 & 24 \\
Hidden layer size & 768 & 1024 \\
Total Parameters & 133M & 414M \\
Batch size & 80 & 144 \\
Optimizer & \multicolumn{2}{c}{AdamW} \\
Learning rate & 1e-4 & 5e-5 \\
Compute resources & \multicolumn{2}{c}{4 $\times$ NVIDIA A100} \\
\hline
\end{tabular}
\caption{\label{tab:korquad_hp} Hyperparameters of LittleBird models submitted to KorQuAD2.0}
\end{table}

\section{How expressive is LittleBird Attention?}
One might think that the linear bias and pack and unpack attention of BiALiBi cannot compete with the flexibility of trainable positional embeddings. We conducted experiments on this as well, but we determined that it was not necessary for the core content of this paper, so the results were simply inserted into the appendix.

We collected attention distributions for each head of layers when feeding held out data to base-size KoELECTRA and Korean LittleBird to investigate the effect of trainable positional embedding and USWAttention with BiALiBi. For KoELECTRA and Korean LittleBird, sequences of length 512 and 1536 were fed, respectively. The collected distributions were converted into probabilities based on query positions and key positions, and the average was then converted to log scale to be plotted. Figure \ref{fig:attention_heatmap} shows 9 selected heads of various layers from the results. The x-axis in each subfigure represents the query position, and the y-axis represents the key position. The brighter the color, the more intense the attention.

Attention patterns between layer heads are comparable due to the distillation of the attention distribution.
In trainable positional embedding, repeating diagonal stripes are a common occurrence. Given the nature of text data, translation-invariance, this is an unusual result that could even be considered noise.
In the case of USWAttention with BiALiBi, on the other hand, attention intensity decreases consistently with distance due to a strong inductive bias from linear bias, and it can be seen that the insufficient part is compensated by pack and unpack attention. LittleBird's Attention module, in particular, seems to have learned key patterns (attention to far-away, attention to near-left or near-right, attention to far-left or far-right, etc.) by distilling from trainable positional embeddings.
Considering the above results, it is clear that LittleBird's attention provides adequate capacity for natural language modeling.

\begin{figure*}
    \begin{subfigure}[b]{1\textwidth}
    \centering
    \includegraphics[width=0.667\textwidth]{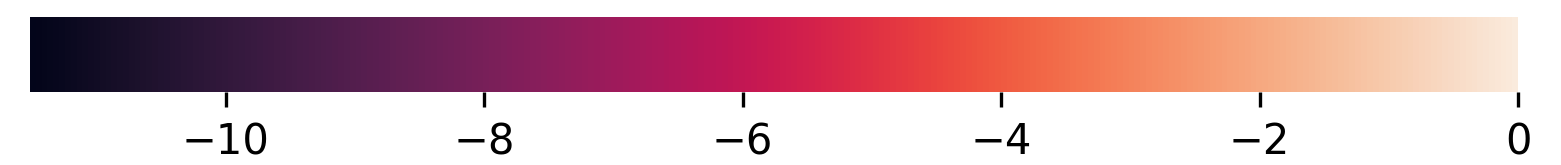}
    \end{subfigure}
    \begin{subfigure}[b]{1\columnwidth}
    \centering
    \includegraphics[width=0.9\columnwidth]{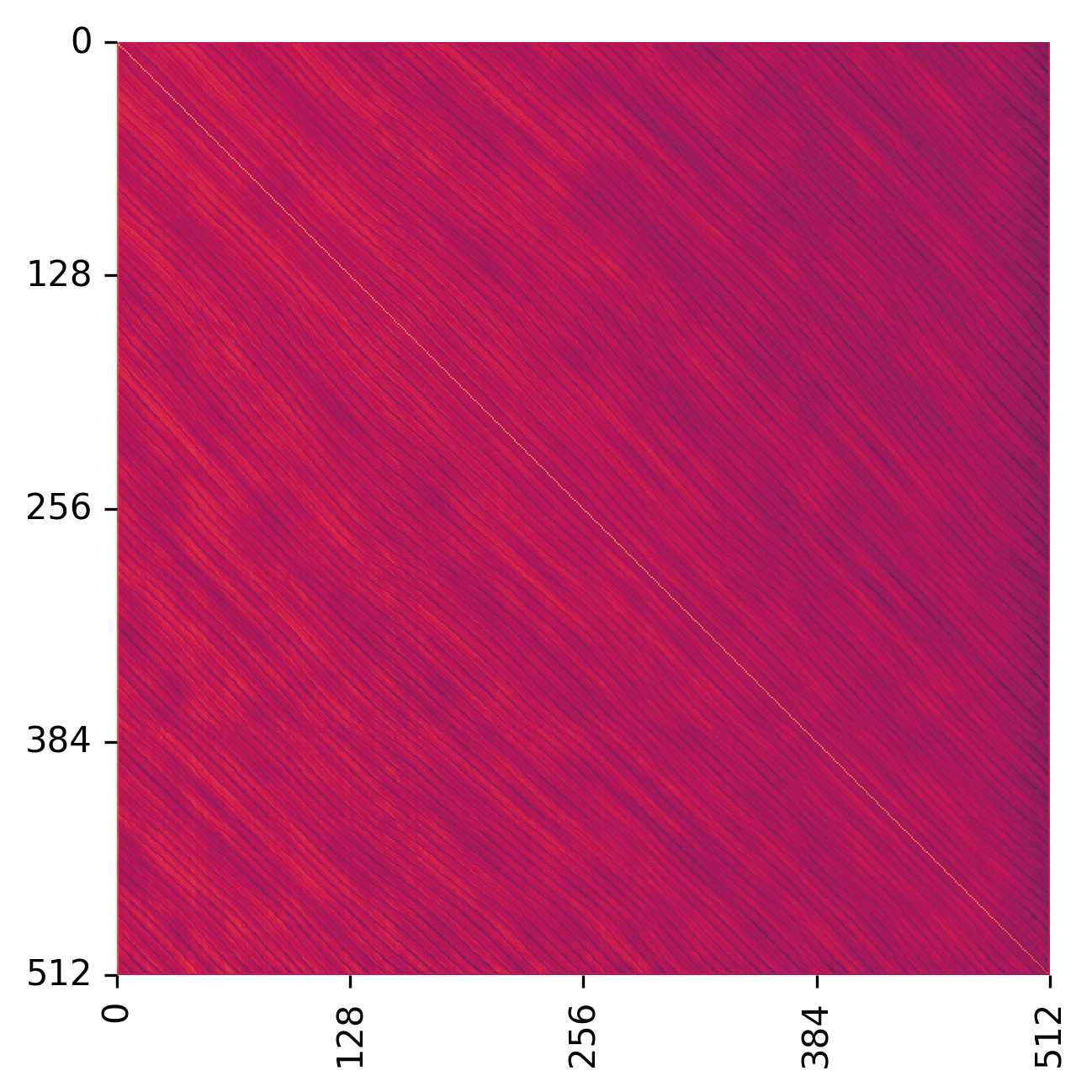}
    \subcaption{ELECTRA, Layer 0 - Head 0}
    \end{subfigure}
    \begin{subfigure}[b]{1\columnwidth}
    \centering
    \includegraphics[width=0.9\columnwidth]{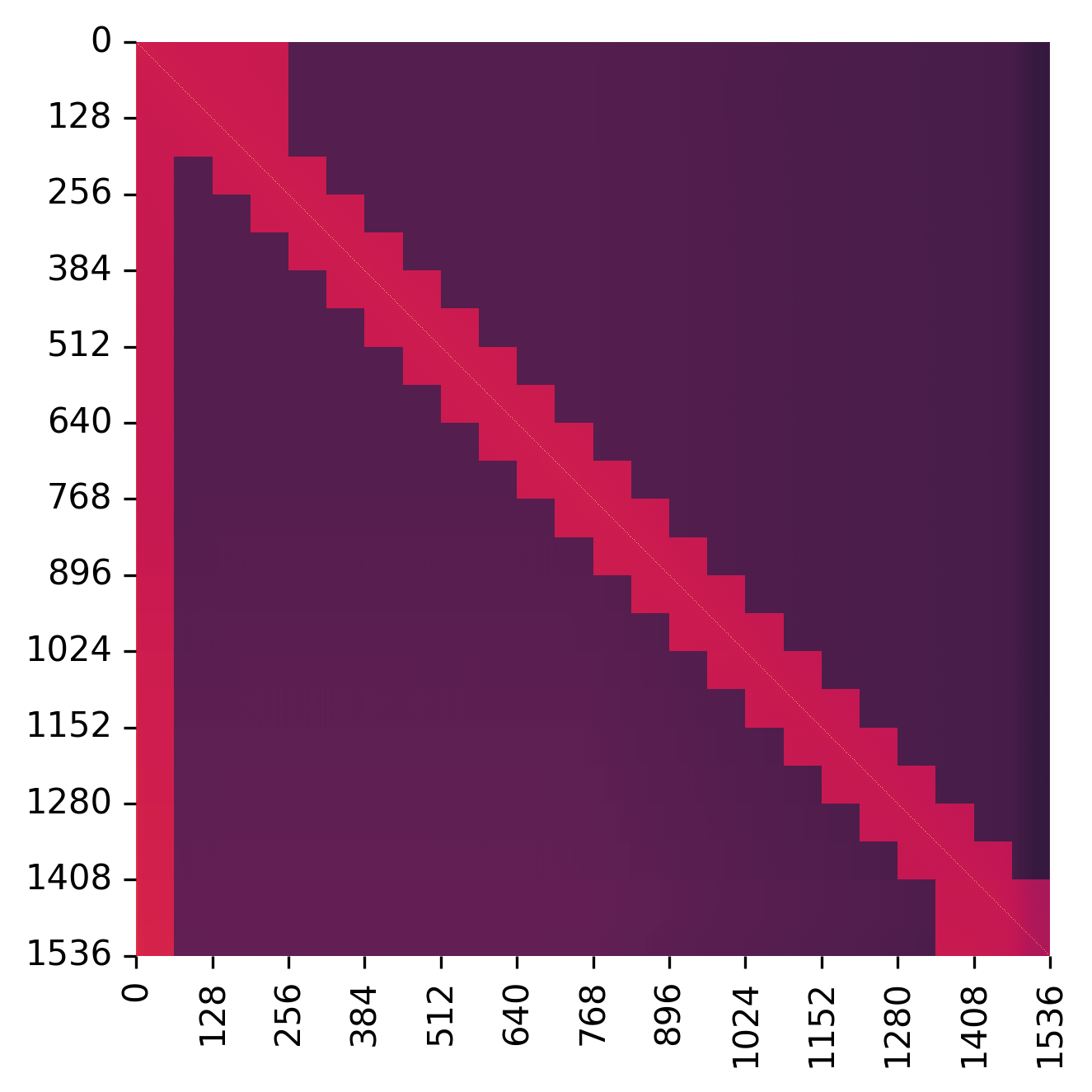}
    \subcaption{LittleBird, Layer 0 - Head 0}
    \end{subfigure}
    \begin{subfigure}[b]{1\columnwidth}
    \centering
    \includegraphics[width=0.9\columnwidth]{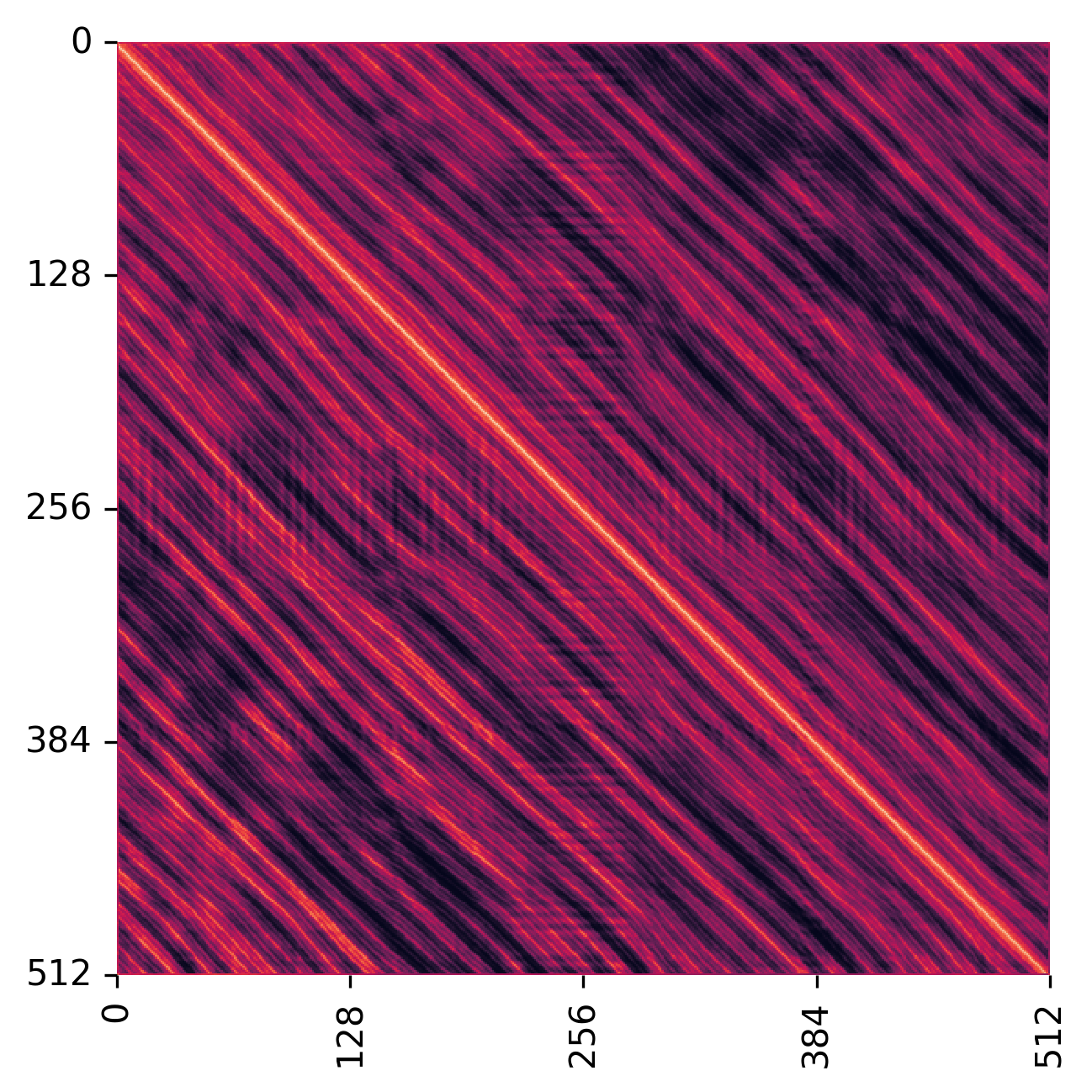}
    \subcaption{ELECTRA, Layer 0 - Head 9}
    \end{subfigure}
    \begin{subfigure}[b]{1\columnwidth}
    \centering
    \includegraphics[width=0.9\columnwidth]{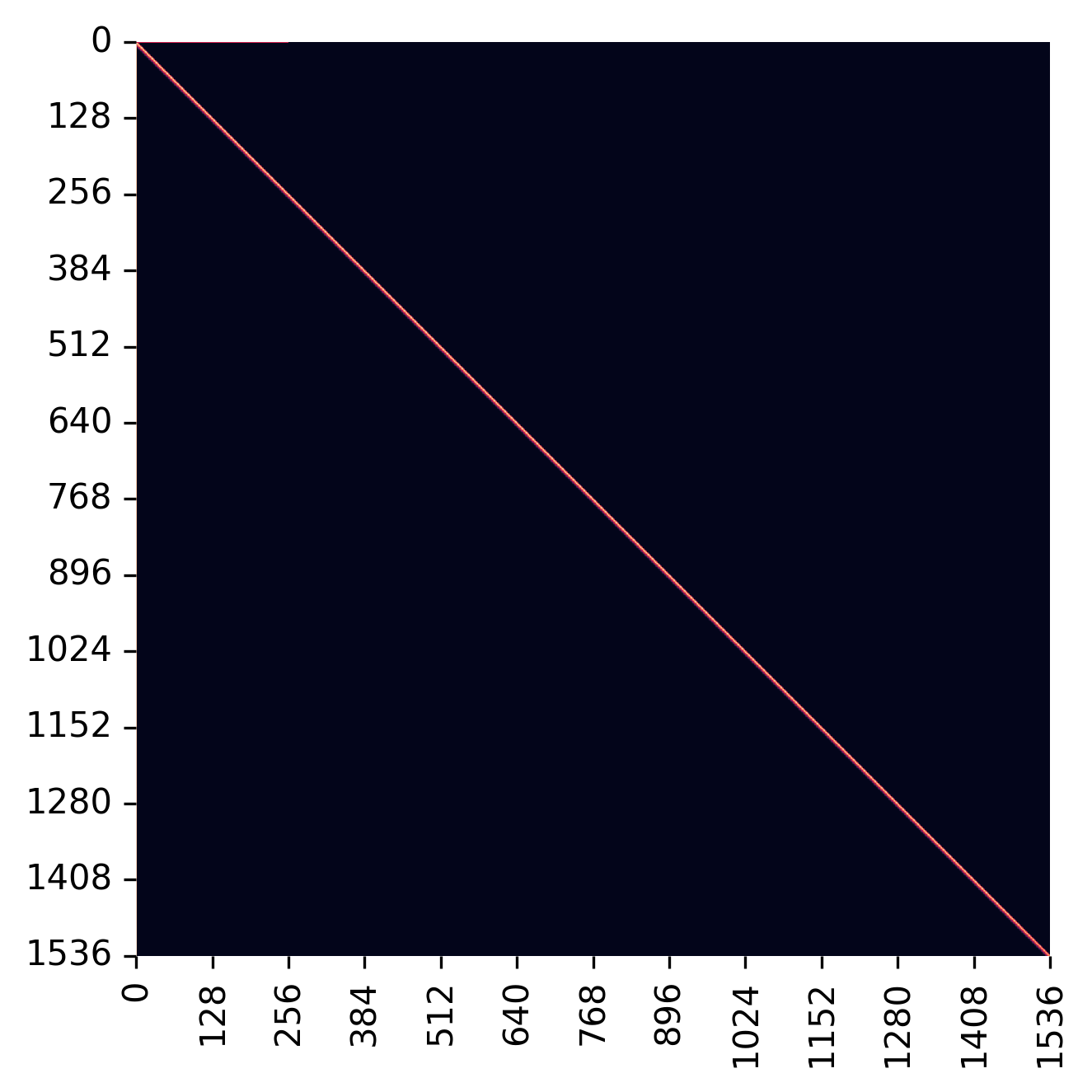}
    \subcaption{LittleBird, Layer 0 - Head 9}
    \end{subfigure}
    \begin{subfigure}[b]{1\columnwidth}
    \centering
    \includegraphics[width=0.9\columnwidth]{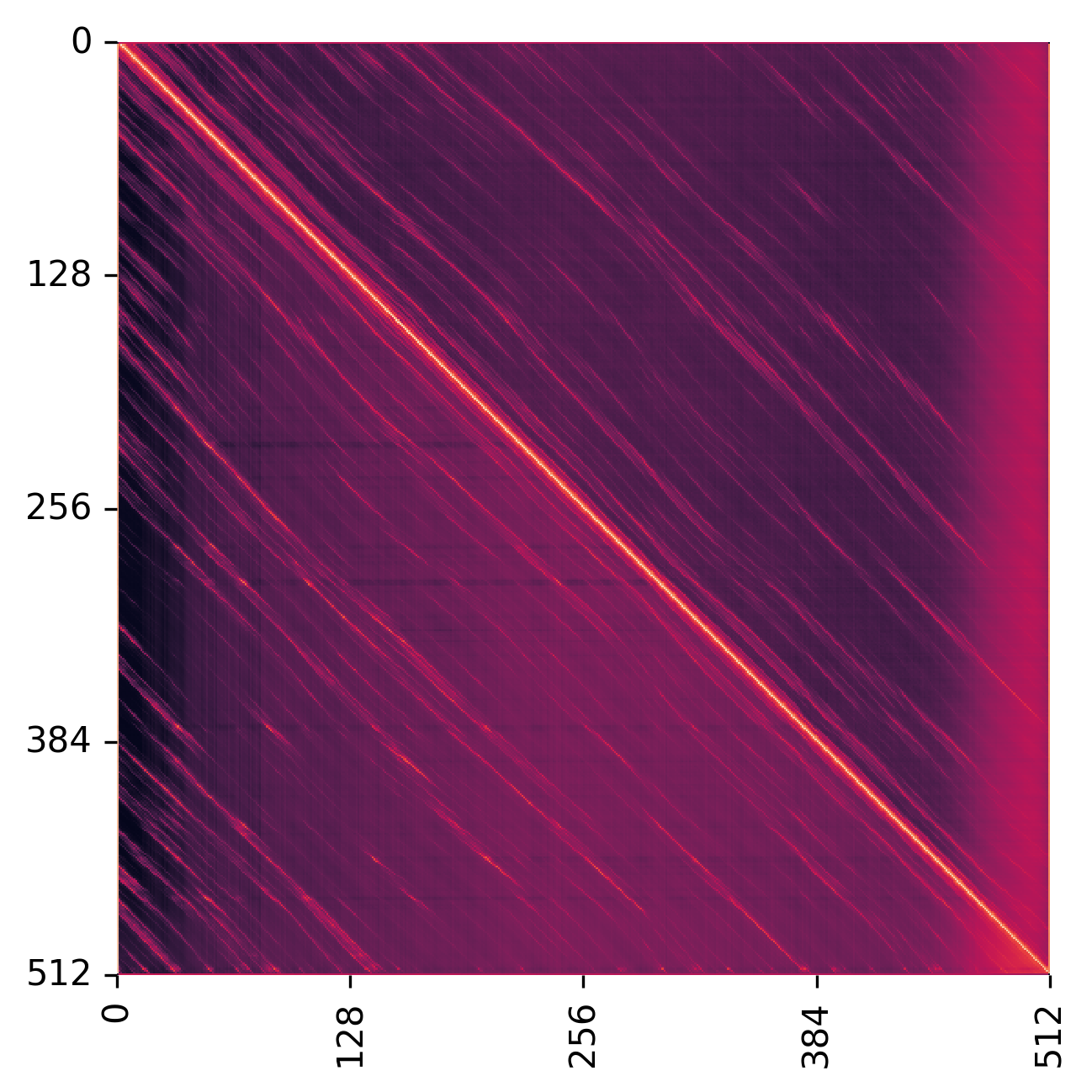}
    \subcaption{ELECTRA, Layer 1 - Head 3}
    \end{subfigure}
    \begin{subfigure}[b]{1\columnwidth}
    \centering
    \includegraphics[width=0.9\columnwidth]{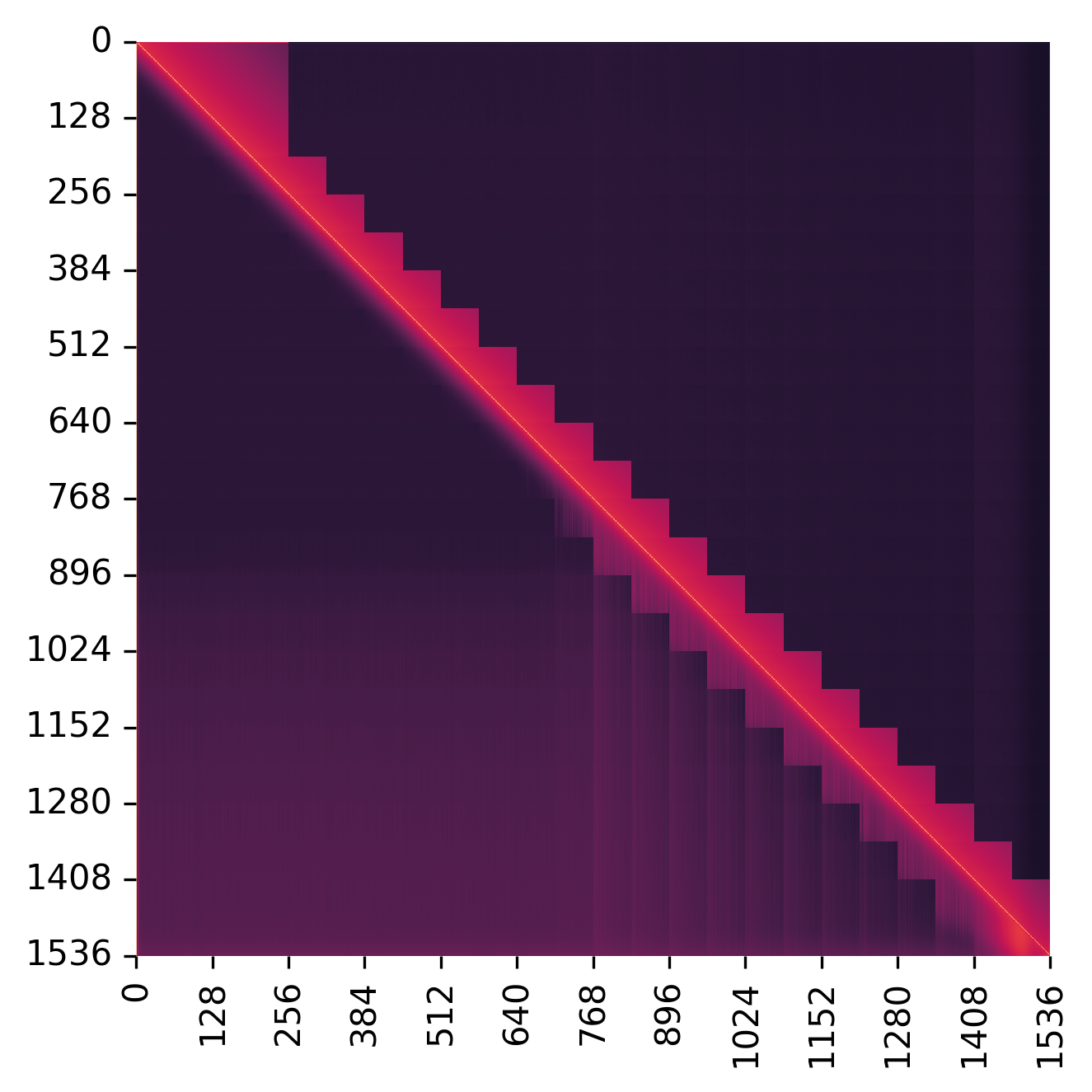}
    \subcaption{LittleBird, Layer 1 - Head 3}
    \end{subfigure}
    \caption{\label{fig:attention_heatmap} Average attention heatmap of ELECTRA and LittleBird}
\end{figure*}
\begin{figure*}\ContinuedFloat
    \begin{subfigure}[b]{1\columnwidth}
    \centering
    \includegraphics[width=0.9\columnwidth]{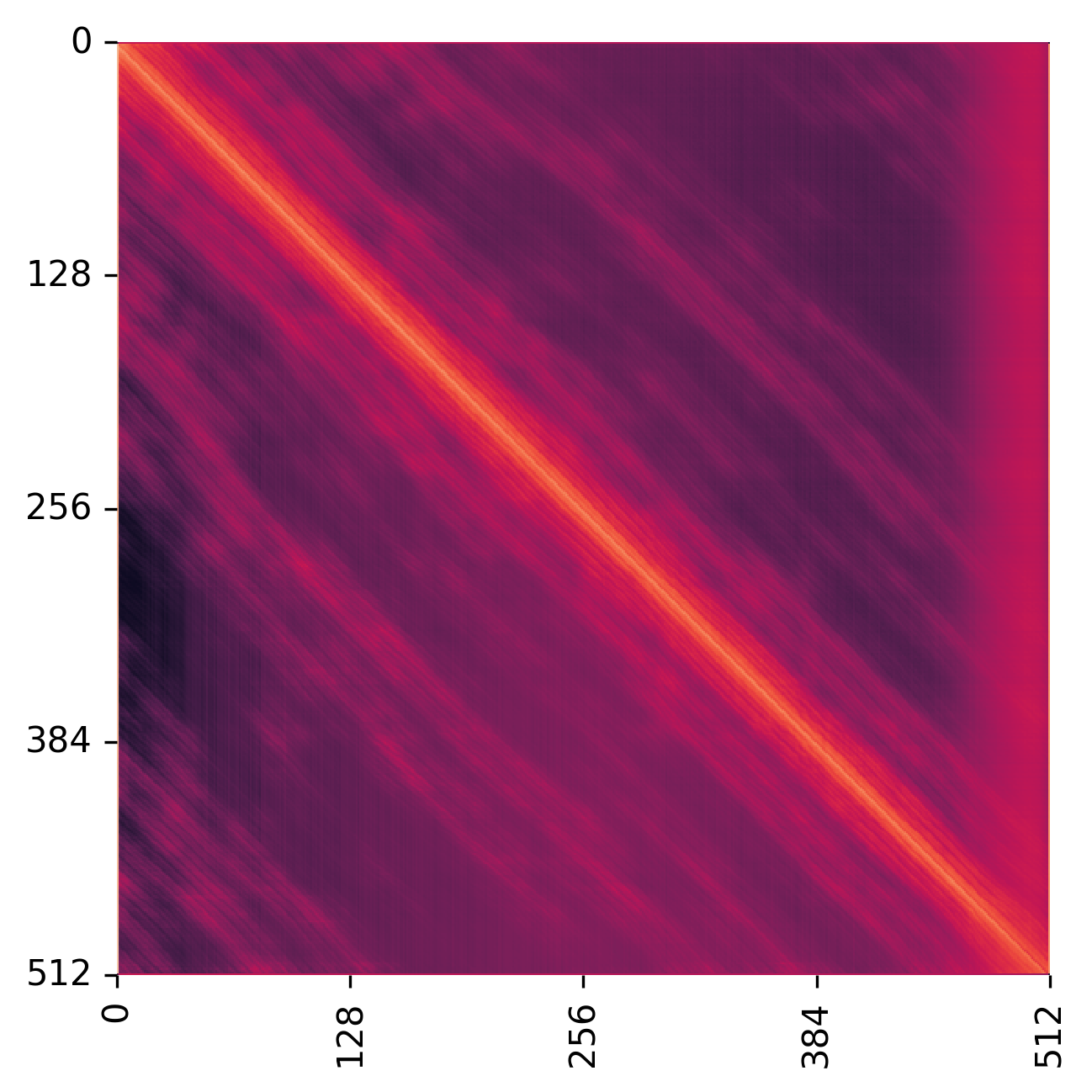}
    \subcaption{ELECTRA, Layer 1 - Head 4}
    \end{subfigure}
    \begin{subfigure}[b]{1\columnwidth}
    \centering
    \includegraphics[width=0.9\columnwidth]{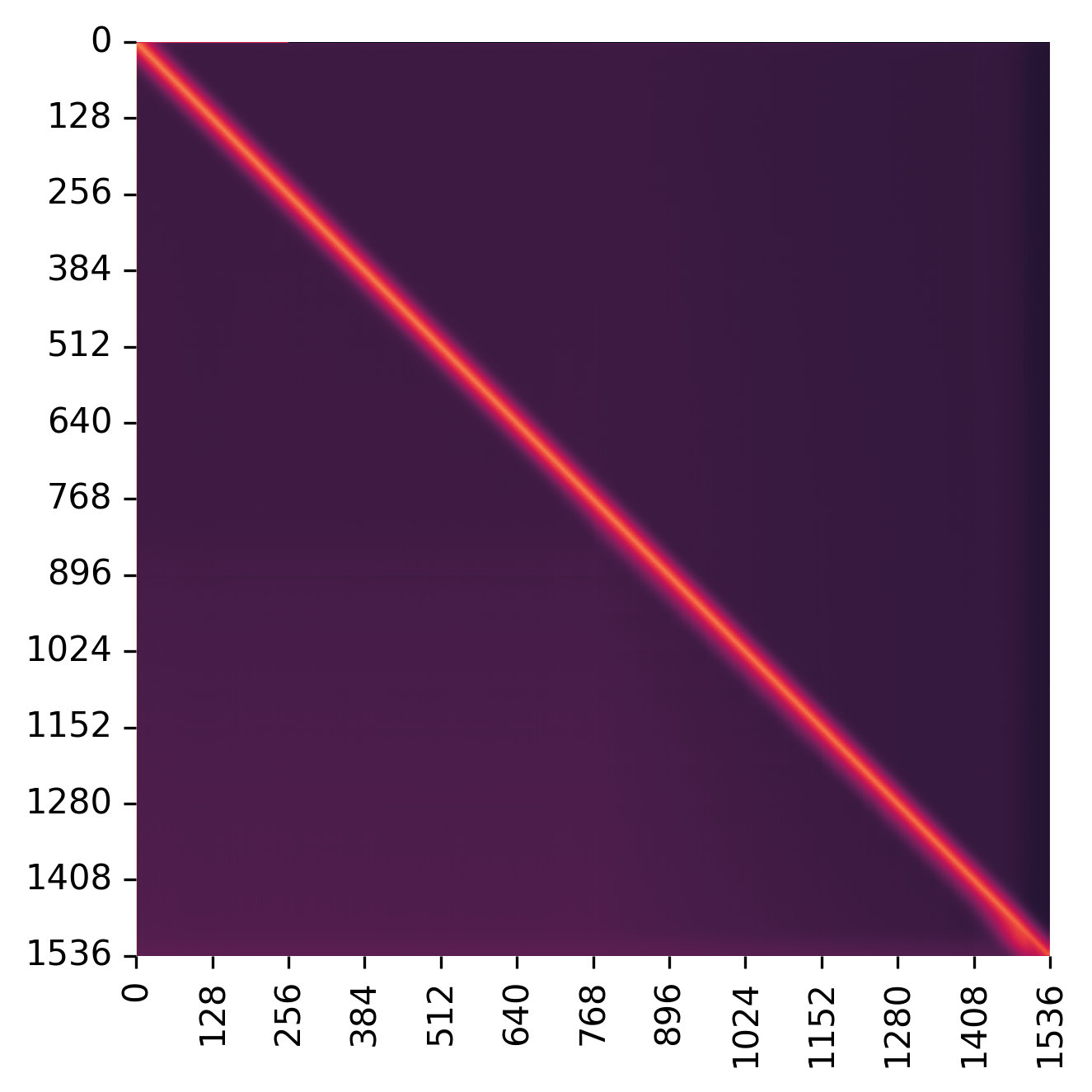}
    \subcaption{LittleBird, Layer 1 - Head 4}
    \end{subfigure}
    \begin{subfigure}[b]{1\columnwidth}
    \centering
    \includegraphics[width=0.9\columnwidth]{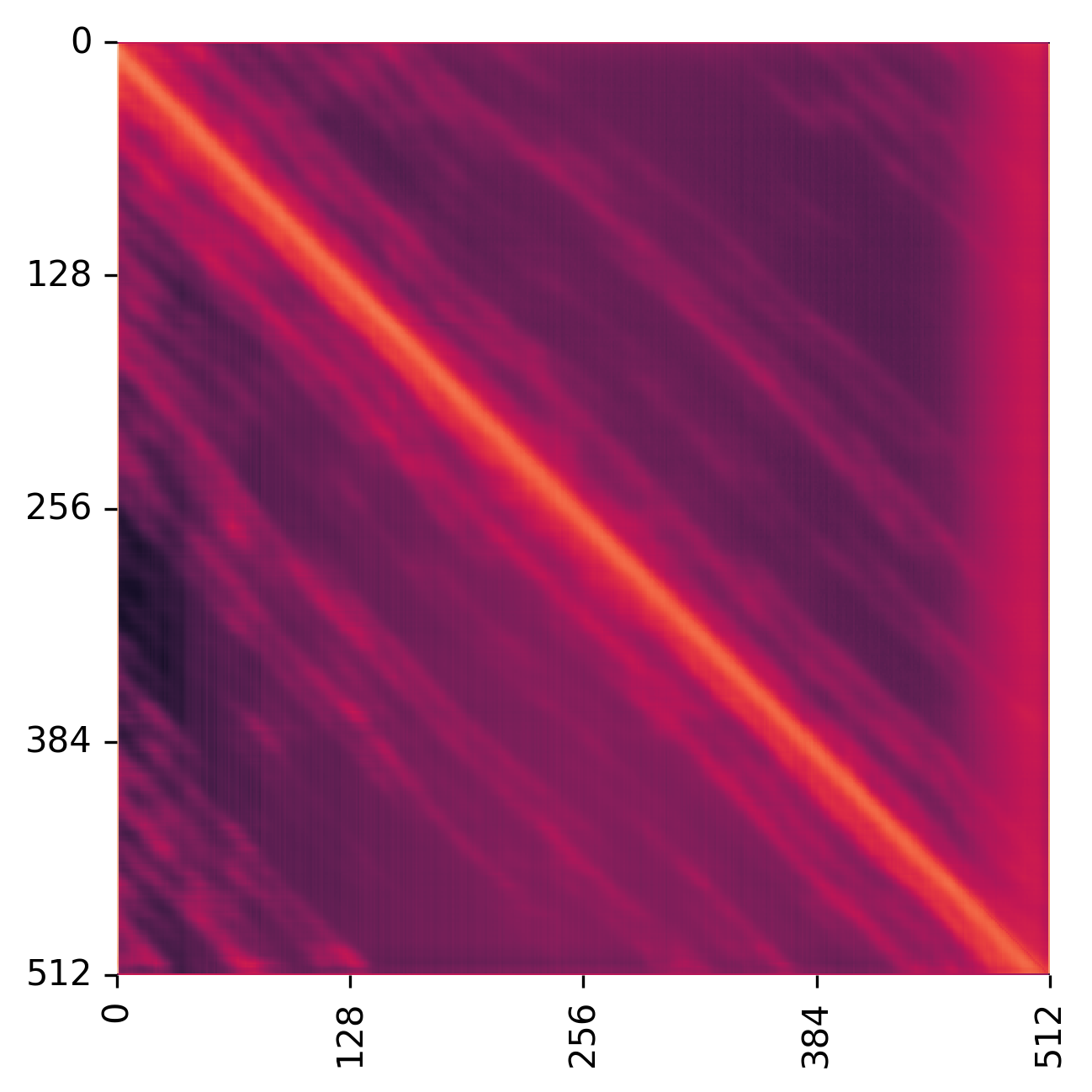}
    \subcaption{ELECTRA, Layer 2 - Head 4}
    \end{subfigure}
    \begin{subfigure}[b]{1\columnwidth}
    \centering
    \includegraphics[width=0.9\columnwidth]{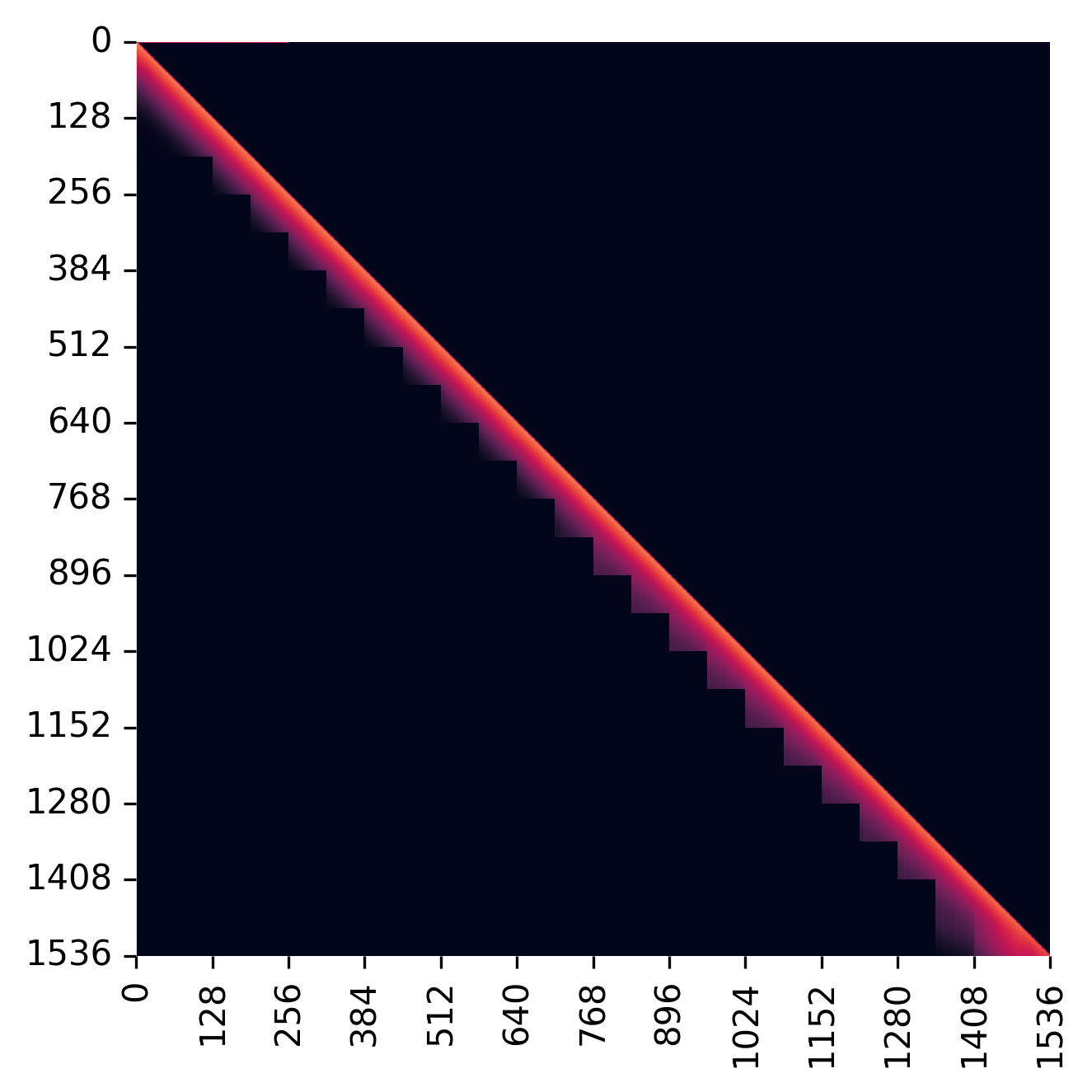}
    \subcaption{LittleBird, Layer 2 - Head 4}
    \end{subfigure}
    \begin{subfigure}[b]{1\columnwidth}
    \centering
    \includegraphics[width=0.9\columnwidth]{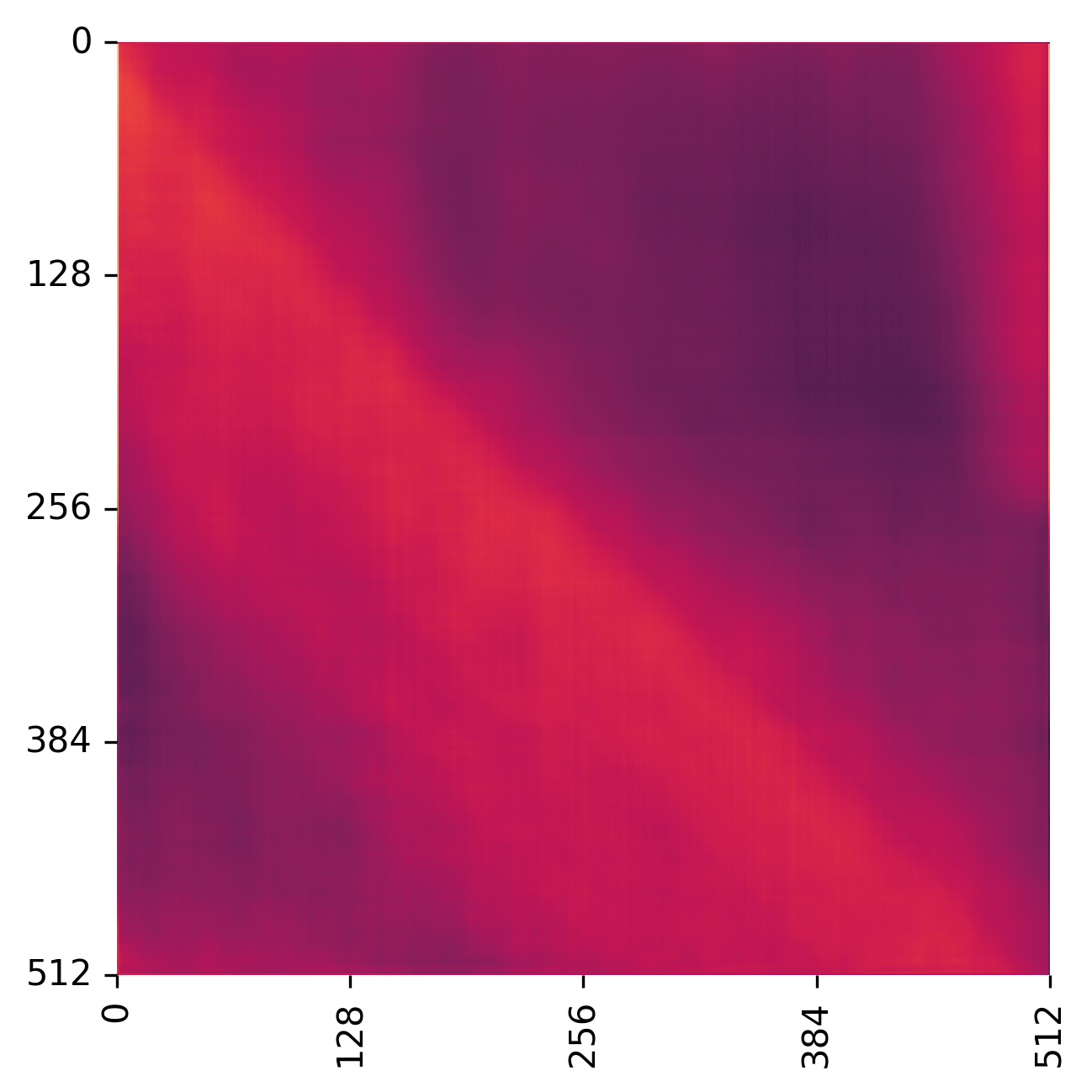}
    \subcaption{ELECTRA, Layer 4 - Head 4}
    \end{subfigure}
    \begin{subfigure}[b]{1\columnwidth}
    \centering
    \includegraphics[width=0.9\columnwidth]{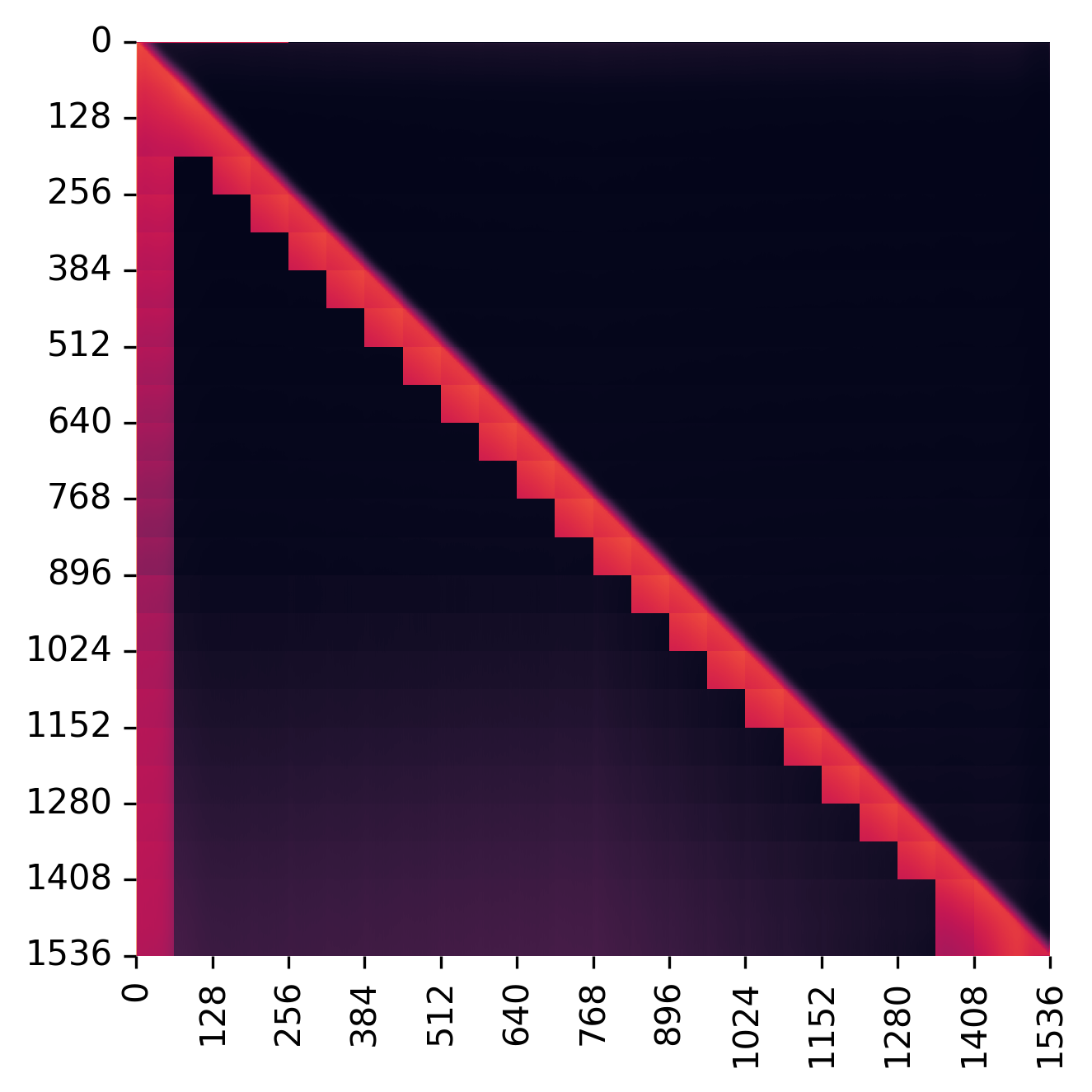}
    \subcaption{LittleBird, Layer 4 - Head 4}
    \end{subfigure}
    \caption{Average attention heatmap of ELECTRA and LittleBird (cont.)}
\end{figure*}
\begin{figure*}\ContinuedFloat
    \begin{subfigure}[b]{1\columnwidth}
    \centering
    \includegraphics[width=0.9\columnwidth]{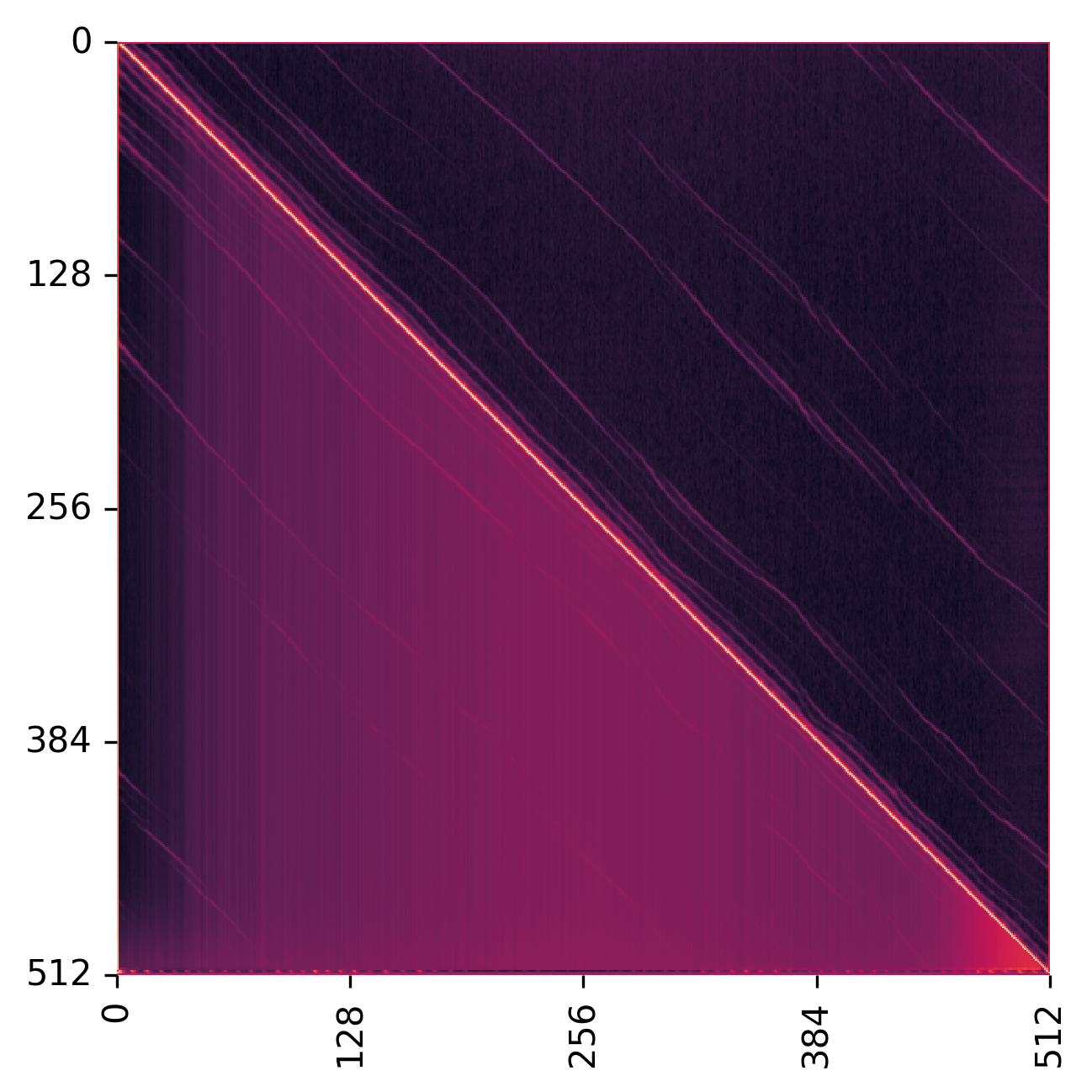}
    \subcaption{ELECTRA, Layer 7 - Head 2}
    \end{subfigure}
    \begin{subfigure}[b]{1\columnwidth}
    \centering
    \includegraphics[width=0.9\columnwidth]{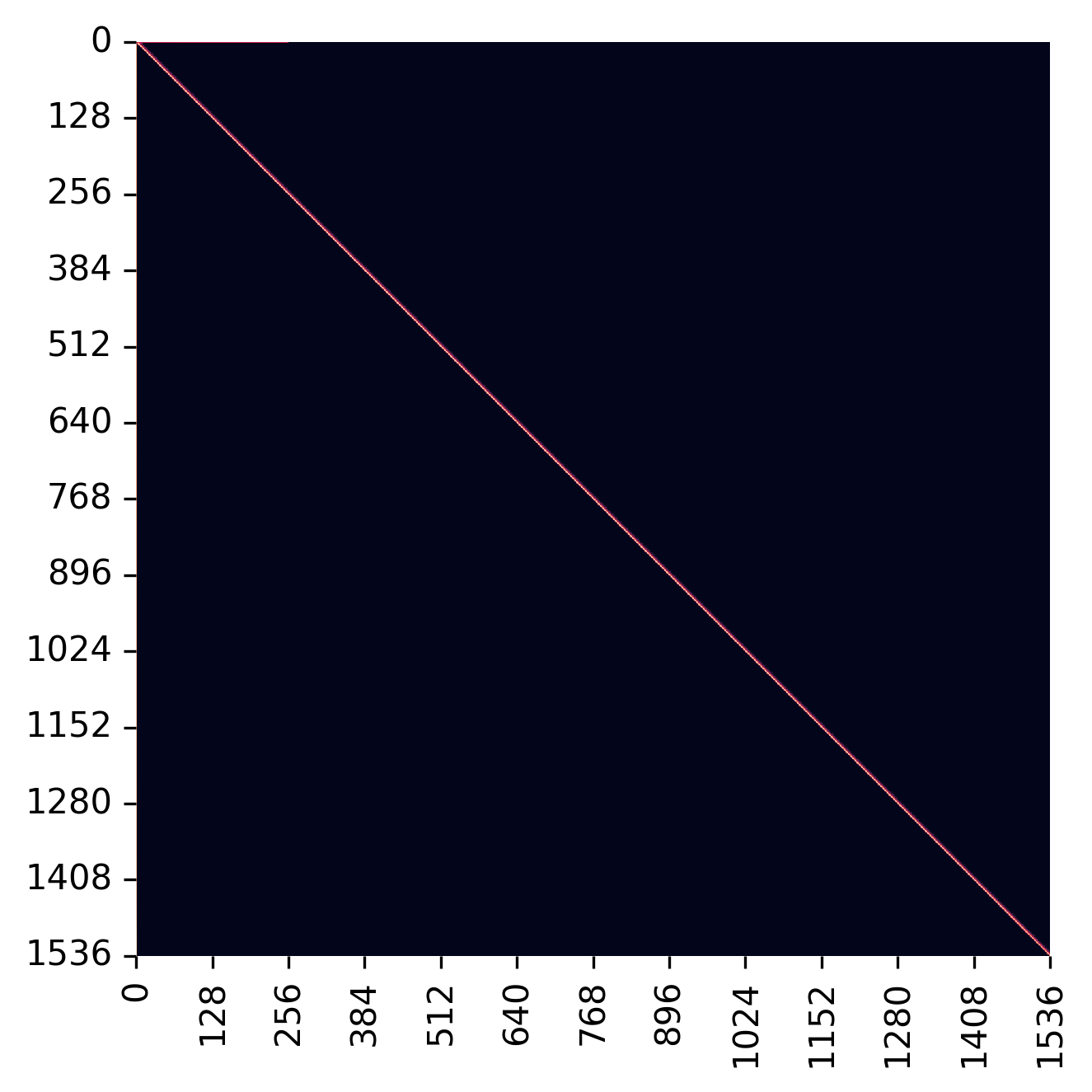}
    \subcaption{LittleBird, Layer 7 - Head 2}
    \end{subfigure}
    \begin{subfigure}[b]{1\columnwidth}
    \centering
    \includegraphics[width=0.9\columnwidth]{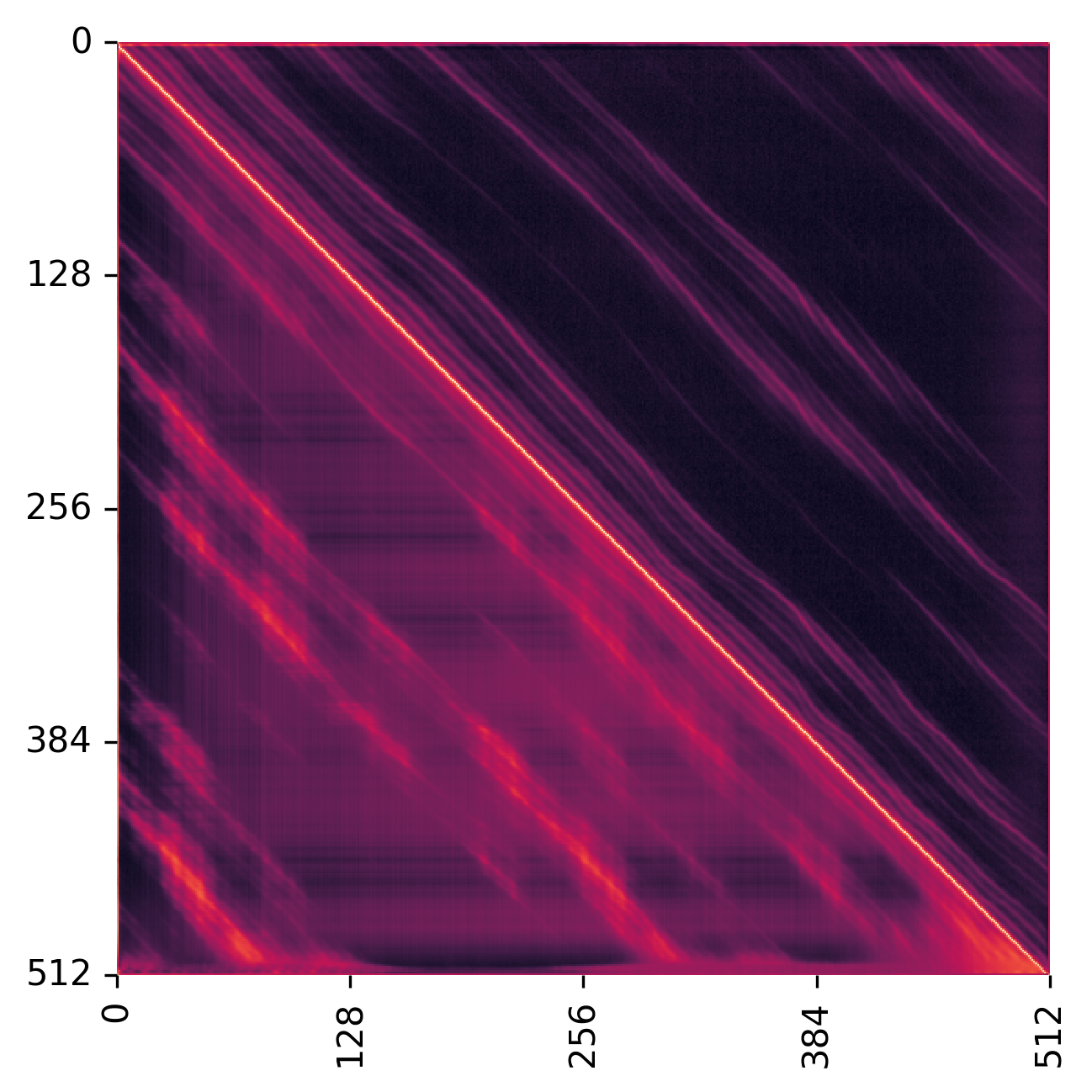}
    \subcaption{ELECTRA, Layer 8 - Head 0}
    \end{subfigure}
    \begin{subfigure}[b]{1\columnwidth}
    \centering
    \includegraphics[width=0.9\columnwidth]{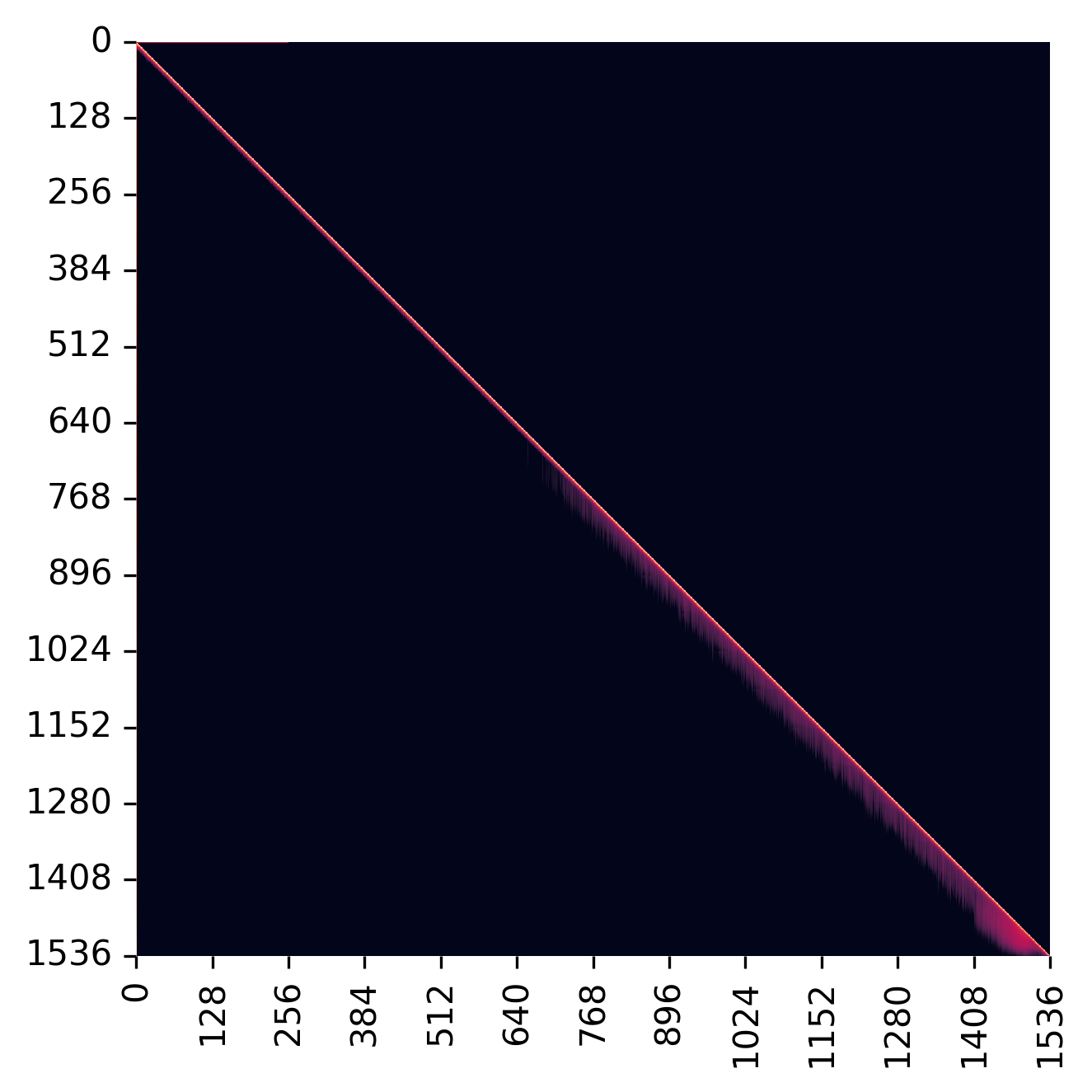}
    \subcaption{LittleBird, Layer 8 - Head 0}
    \end{subfigure}
    \begin{subfigure}[b]{1\columnwidth}
    \centering
    \includegraphics[width=0.9\columnwidth]{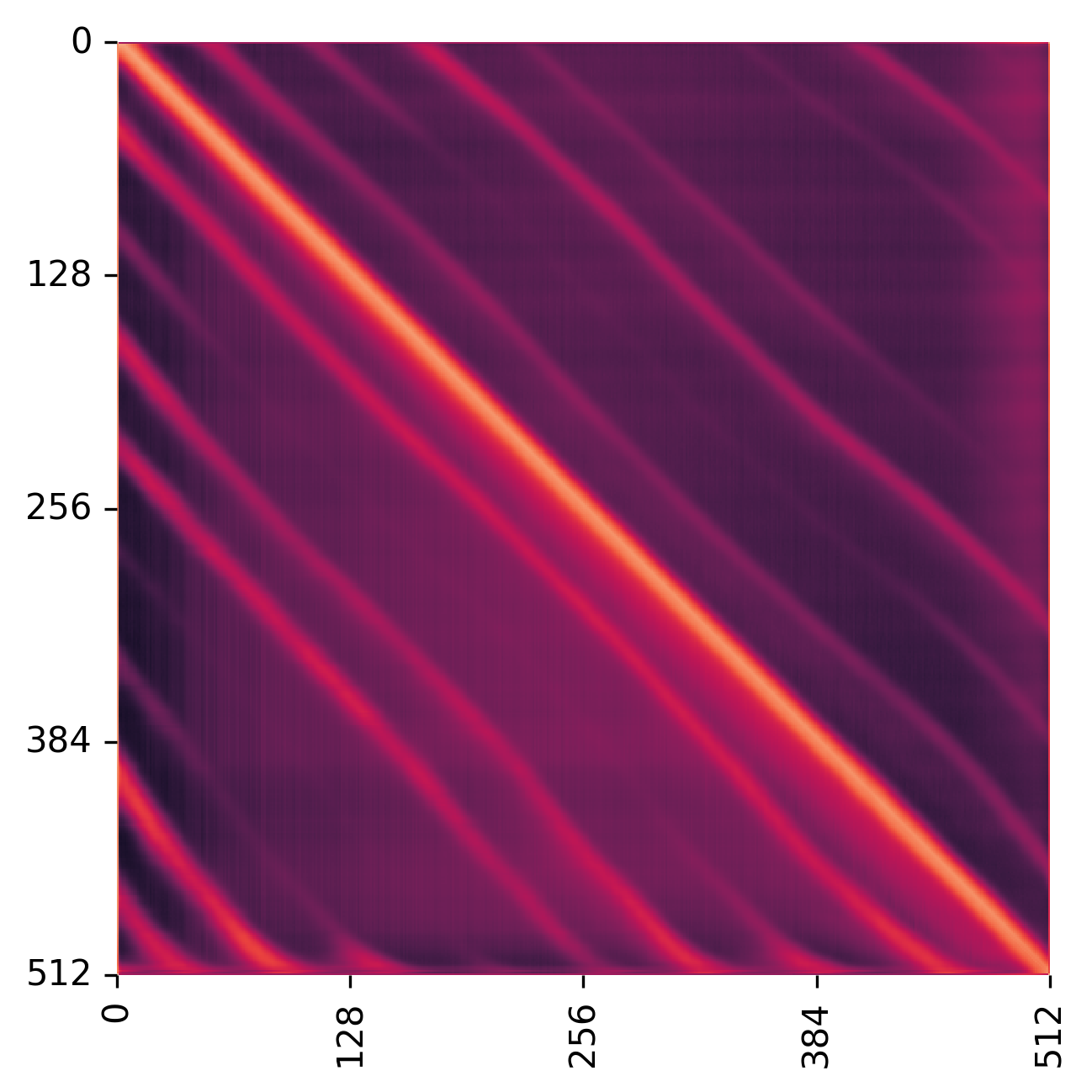}
    \subcaption{ELECTRA, Layer 11 - Head 8}
    \end{subfigure}
    \begin{subfigure}[b]{1\columnwidth}
    \centering
    \includegraphics[width=0.9\columnwidth]{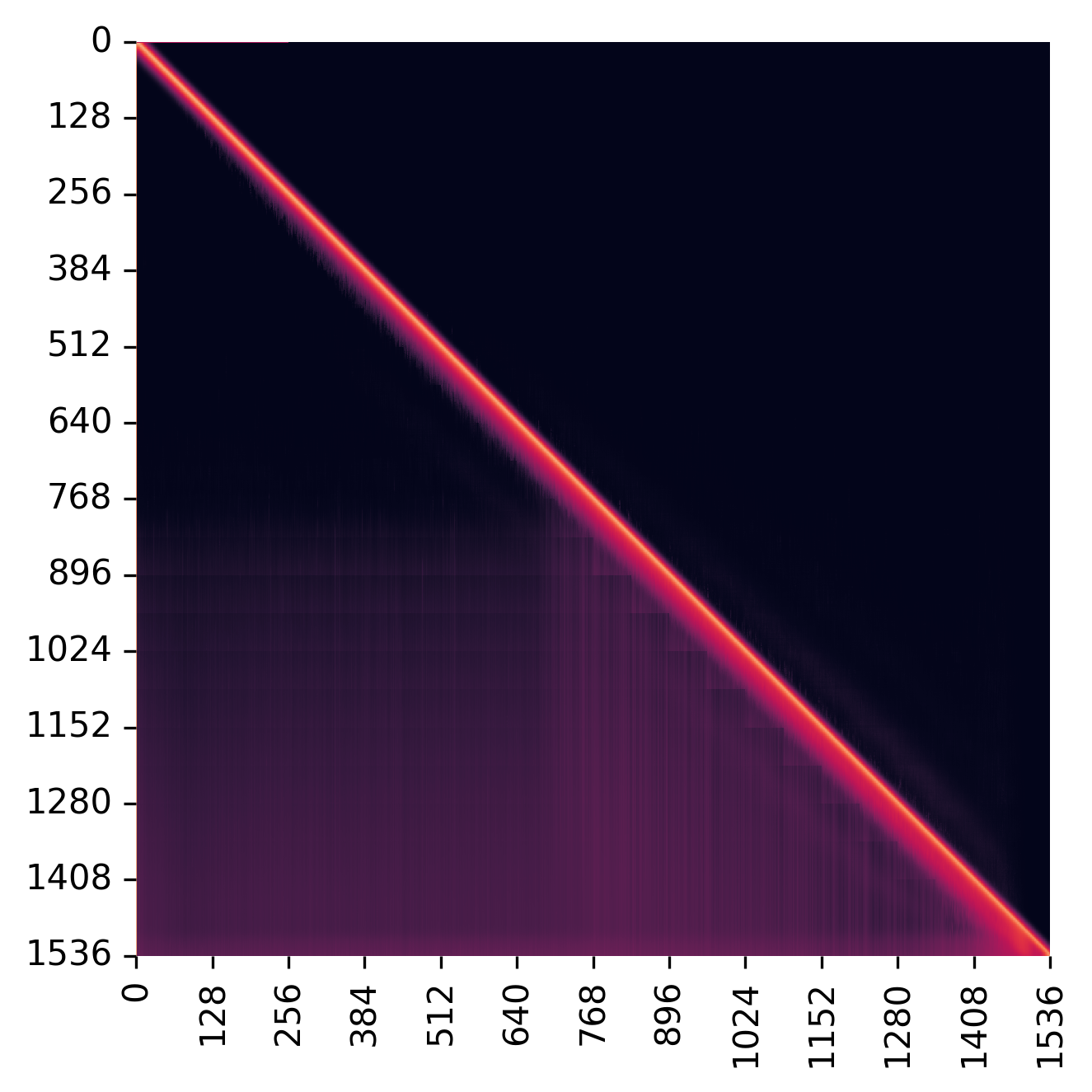}
    \subcaption{LittleBird, Layer 11 - Head 8}
    \end{subfigure}
    \caption{Average attention heatmap of ELECTRA and LittleBird (cont.)}
\end{figure*}


\end{document}